\documentclass{nle_altered}

\usepackage{adjustbox}

\usepackage{tikz}
\usetikzlibrary{positioning}
\usepackage{amsmath}
\usepackage{url}
\makeatletter
\newcommand{\@BIBLABEL}{\@emptybiblabel}
\newcommand{\@emptybiblabel}[1]{}
\let\O@argtabularcr\@argtabularcr
\def\O@xtabularcr{\@ifnextchar[\O@argtabularcr{\ifnum 0=`{\fi}\cr}}
\let\O@tabacol\@tabacol
\let\O@tabclassiv\@tabclassiv
\let\O@tabclassz\@tabclassz
\let\O@tabarray\@tabarray
\def\author@tabular{\authorsize\def\@halignto{}\@authortable}
\let\endauthor@tabular=\endtabular
\def\author@tabcrone{{\ifnum0=`}\fi\O@xtabularcr\authorsize
 \let\\=\author@tabcrtwo\ignorespaces}
\def\author@tabcrtwo{{\ifnum0=`}\fi\O@xtabularcr[-3\p@]\affilsize\itshape
 \let\\=\author@tabcrtwo\ignorespaces}
\def\@authortable{\leavevmode \hbox \bgroup $\let\@acol\O@tabacol
 \let\@classz\O@tabclassz \let\@classiv\O@tabclassiv
 \let\\=\author@tabcrone \ignorespaces \O@tabarray}
\makeatother
\usepackage[hidelinks]{hyperref}

\usepackage{apacite}
\usepackage{graphicx}
\usepackage{tabularx}
\usepackage{enumitem}
\DeclareMathOperator{\Attention}{Attention}
\DeclareMathOperator{\softmax}{softmax}

\title[Open-type relation argument extraction]
      {Neural architectures for\\ open-type relation argument extraction}

\author[Open-Type Relation Argument Extraction]
{Benjamin Roth$^\ast$, Costanza Conforti$^\dagger$, Nina Poerner$^\ast$,\\Sanjeev Karn$^\ast$, Hinrich Sch{\"u}tze$^\ast$ \\\\ $^\ast$Center for Information and Language Processing, Ludwig Maxilian University of Munich\\$^\dagger$Language Technology Laboratory, University of Cambridge\\\url{beroth@cis.uni-muenchen.de}, \url{inquiries@cislmu.org}}

\pagerange{\pageref{firstpage}--\pageref{lastpage}}
\pubyear{2018}

\let\vec\mathbf

\let\cite\shortcite
\let\citeA\shortciteA

\newcounter{notecounter}

\newcommand{\enoteson}{\long\gdef\enote##1##2{{
\stepcounter{notecounter}
{\large\bf
\hspace{1cm}\arabic{notecounter} $<<<$ ##1: ##2
$>>>$\hspace{1cm}}}}}
\enoteson

\begin{document}

\label{firstpage}
\maketitle

\begin{abstract}
In this work, we introduce the task of \emph{Open-Type Relation
  Argument Extraction (ORAE)}:
Given a corpus, a query entity $Q$ and a knowledge base relation (e.g.,
``$Q$ authored notable work with title $X$''),
the model has to extract
an argument of non-standard entity type (entities that cannot be extracted by a standard named entity tagger, e.g., $X$: the title of
a book or a work of art) from the
corpus. 

We develop and compare a wide range of neural models for this task yielding large improvements over a strong baseline obtained with a neural question answering system. The impact of different sentence encoding architectures and answer extraction methods is systematically compared. An encoder based on gated recurrent units combined with a conditional random fields tagger yields the best results.
We release a data set to train and evaluate ORAE, based on WikiData and obtained by distant supervision.
\end{abstract}

\section{Introduction}
Systems for turning unstructured information from textual corpora (such as Wikipedia and newspaper corpora) into structured representations are crucial tools for harnessing the vast amounts of data available on-line.
Automatic detection of relations in text allows humans to search and find relevant facts about entities, and it allows for further processing and aggregation of relational information.
A prototypical user for such a system would be, e.g., an analyst who is interested in facts about a specfic organization or person, or a social scientist who is interested in aggregating facts over time for trend detection.

Entity-driven relation extraction is the problem of
identifying relevant facts for a query entity $Q$
(e.g., $Q$ = ``Steve Jackson'')
in a large corpus according to
a pre-defined relational schema that defines relations such as
``$Q$ authored notable work with title $X$''.  Systems solving
this task are often complex pipelines containing modules for
information retrieval, linguistic pre-processing and
relation classification (cf. \citeA{surdeanu2013overview}). 

While the main focus in relation extraction has previously been on
\emph{relation classification} (i.e., predicting whether a relation holds
between two given arguments), quantitative analysis
has repeatedly shown that \emph{argument identification}
(often performed by carefully
engineered submodules)
has at least as big of an impact on
end-to-end results \cite{pink2014analysing,roth2015effective}. Moreover, in previous benchmarks \cite{surdeanu2013overview, zhang2017position},
relations have been selected such that the vast majority of arguments are
of standard types (e.g., \textit{person}, \textit{location},
\textit{organization}) and can be detected by a named entity
recognizer.
Even for standard named
entity types,
argument identification is hard for
complex cases like nested named entities because
different levels of granularity are relevant for
different relations.

Consider, for example, the following two named entity tagging errors:
\begin{itemize}
 \item \emph{[Popular Kabul]$_{ORG}$ lawmaker [Ramazan Bashardost]$_{PER}$ , who camps out in a tent near parliament ...}
 \item \emph{[Haig]$_{PER}$ attended the [US Army]$_{ORG}$ academy at [West Point]$_{LOC}$ ...}
\end{itemize}
In the above example, a pipelined system which relies on the tagging output cannot extract \emph{Kabul} as the \emph{city-of-residence} for the query \emph{Ramazan Bashardost}. It can also not extract \emph{US Army academy at West Point} as the \emph{school-attended} for the query \emph{Haig}, even though the relation is expressed explicitly by the verb \emph{attended}. Argument identification 
of nonstandard types (e.g., a title of a book or a
work of art), which is the focus of this work, is even more challenging.

Comparison of end-to-end relation extraction systems,
as in the Knowledge Base Population (KBP) English Slot Filling shared task \cite{surdeanu2013overview,angeli2014combining},
indicates that recall is the most difficult metric to optimize in
entity-driven relation extraction. Further analysis
\cite{pink2014analysing} 
showed that named entity tagging is, after relation
prediction, the main bottleneck accounting for roughly 30\%
of the missing recall. It is also worth noting that tagging or matching errors may harm twice: once for missing the correct answer, and secondly for returning an incorrect answer span.

The key motivation for our research is that \emph{identification of the query
  entity} is relatively easy and causes few errors: string
match and expansion heuristics using information retrieval methods work well
and need not rely on entity tagging. In contrast,
\emph{identification of the slot filler} is hard,
especially if a diverse range of entity types is
considered. 
Consequently, \emph{we give the relation prediction model full freedom to select a slot filler
from all possible sub-sequences of retrieved query contexts.} 

Based on this motivation, we define the task of
\emph{Open-Type Relation Argument Extraction (ORAE)}, a more general form of
entity-driven relation classification.
In contrast to the standard setting (which has been the focus of KBP),
the key novelty of ORAE is that 
slot fillers of any type are admissible; they are not restricted to the
standard entity types like \emph{person} and
\emph{location}.
Broadening the definition of types at the same
time allows us to broaden the definition of relations and
we can handle relations that pose difficulty for standard
relation classification.

Most slot filling methods make heavy use of named entity
recognition \cite{zhang2016stanford}, but named entity recognizers address only 
pre-defined types (for which there is training data with
annotated entities).
Non-standard types cannot be recognized without special
engineering (e.g., compiling lists of entities or writing
regular expressions).
To address this, we propose a set of new
relation argument extraction methods in this article that
do not require a named entity recognizer.

In summary, this article makes the following contributions:
\begin{itemize}
 \item The formulation and motivation of Open-Type Relation
   Argument Extraction (ORAE) as a problem in
   information extraction, and a novel dataset for Wikidata relations that contain an argument that is of non-standard type.
 \item A range of different neural network architectures
   for solving ORAE and their evaluation in extensive experiments:
 \begin{itemize}
  \item We compare different neural architectures (\emph{encoders}) for computing a sentence representation suitable for argument extraction. The proposed encoders are based on convolutional networks \cite{collobert2011natural}, recurrent networks \cite{chung2014empirical}, and self-attention \cite{vaswani2017attention}.
  \item We compare different neural architectures (\emph{extractors}) for extracting answers from this sentence representation. The proposed extractors are based on pointer models \cite{DBLP:conf/nips/VinyalsFJ15}, linear chain conditional random fields \cite{lafferty2001crf,lample2016neural}, and table filling \cite{miwa2014modeling}.
 \end{itemize}
\end{itemize}

\section{Encoding and extraction architectures}

A big class of errors in end-to-end relation extraction
systems are missing or inexact named entity tags and, in a
pipelined model, lost recall cannot be regained \cite{pink2014analysing,roth2015effective}. The
models we propose aim at overcoming this problem by
skipping the named entity recognition step
altogether, and instead predicting a slot filler (or none) for
query entities and the relations of interest.
Our models do not perform 
a separate task of entity
recognition; but of course
they have to do entity recognition implicitly since
extracting a correct slot filler requires correct assessment
of its type and correct assessment of the type of the query entity.
The aim of this work is to develop models that predict knowledge graph relations for concepts that have non-standard type in a query-driven setup, and to explore a wide range of possible solutions to this problem.

\begin{figure}
\includegraphics[width=\textwidth]{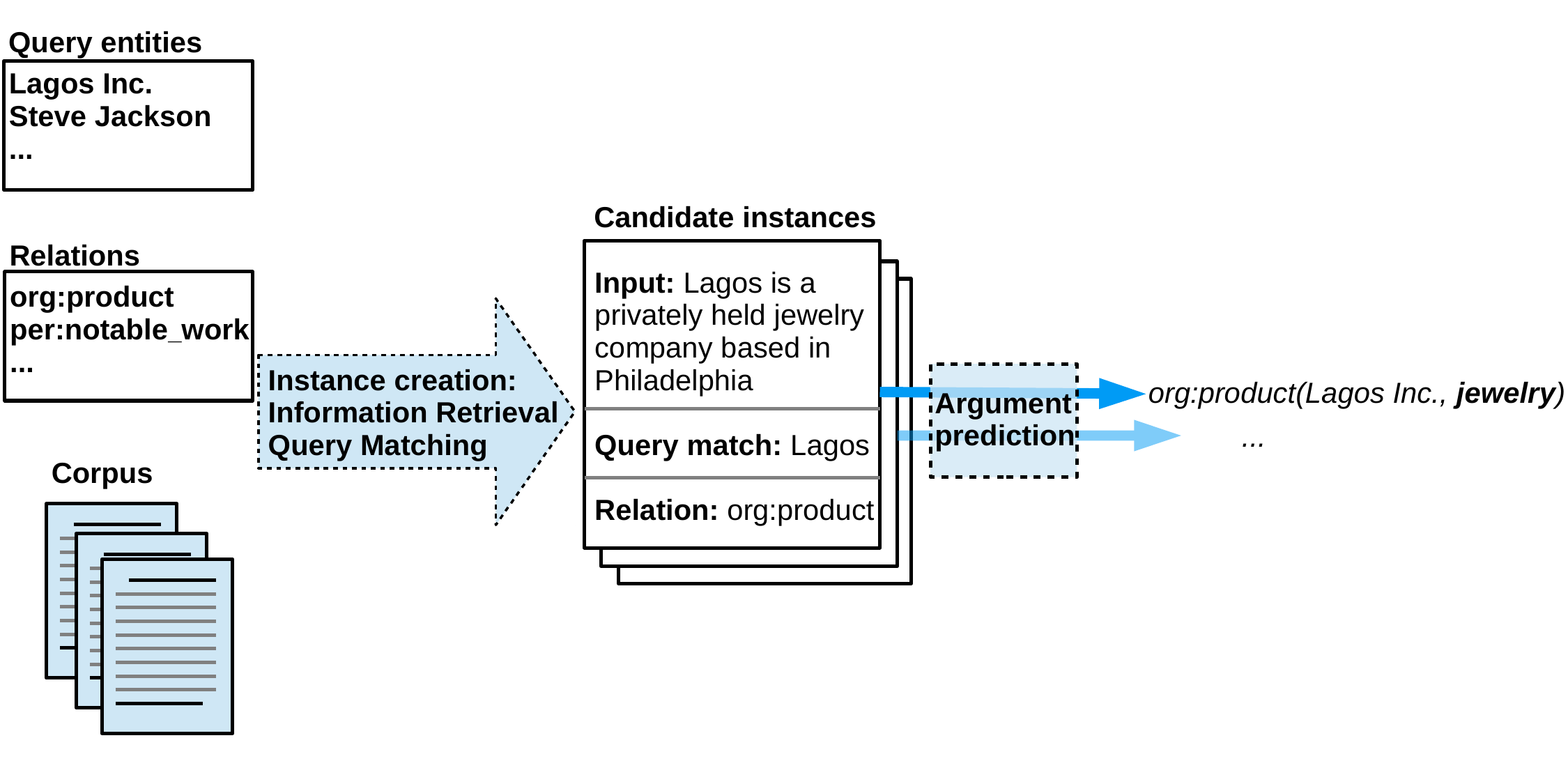}
	\caption{Schematic overview of a query-driven knowledge base population system. The focus of this work is developing an argument prediction component that can extract non-standard entities.}
	\label{fig:big_picture}
\end{figure}

Figure \ref{fig:big_picture} shows the general setup in which our argument prediction models can be applied. The practical scenario is one where 
a user seeks to extract
relational information 
from a large text corpus for a list of relevant query entities and relations (depending on the query entity type, \citeA{surdeanu2013overview}). 
We call this scenario query-driven KBP.
In query-driven KBP, input to the argument prediction model is a context that has been provided by the retrieval system for the relevant query entity, for example:
\begin{itemize}
 \item \textbf{Query:} \emph{``Alexander Haig''}
 \item \textbf{Context:} \emph{``Haig attended the US army academy at Westpoint.''}
\end{itemize}

The relation of interest is also provided to the model (if there are several possible relations for a query type, several instances are created). 
In traditional approaches to query-driven KBP, the query and a second potential argument is marked by named-entity tagging, and a simple classification prediction has to be made for all potential relations, for example:

\begin{itemize}
 \item \emph{``[Haig]$_{Query}$ attended the [US army]$_{Answer}$ academy at Westpoint .''}\\ \verb=works-for= $\Rightarrow$ Yes/No?
 \item \emph{``[Haig]$_{Query}$ attended the [US army]$_{Answer}$ academy at Westpoint .''}\\ \verb=school-attended= $\Rightarrow$ Yes/No?
 \item \emph{``[Haig]$_{Query}$ attended the US army academy at [Westpoint]$_{Answer}$ .''}\\ \verb=born-in= $\Rightarrow$ Yes/No?
 \item ...
\end{itemize}

In our ORAE approach, the answer has to be identified simultaneously with deciding whether the relation holds or not. 

\begin{itemize}
 \item \emph{``[Haig]$_{Query}$ attended the US army academy at Westpoint .''}\\ \verb=works-for= $\Rightarrow$ Answer?\\ 
\verb=school-attended= $\Rightarrow$ Answer?\\ 
\verb=born-in= $\Rightarrow$ Answer?\\
...
\end{itemize}

We conceptually break our models for argument prediction down into three components:
\begin{itemize}
\item \textbf{Lookup layer}: Representation of the context
  sentence. We use the same input representation throughout
our experiments.
\item \textbf{Encoder}: Layers that compute a representation for every position in the sentence, combining information from other positions.
 \item \textbf{Extractor}: Last part of the architecture; it computes the extracted answer as the output.
\end{itemize}
\tikzstyle{label}=[minimum height=5mm, font=\small, inner sep=2pt]
\tikzstyle{box}=[draw, rectangle, label, inner sep=2pt]
\tikzstyle{embedding}=[box, minimum width=4.4cm, label]
\tikzstyle{layer}=[box, minimum width=1.5cm, minimum height=3mm, label]
\tikzstyle{arrow}=[->, rounded corners]
\tikzstyle{elabel}=[font=\bfseries\small, minimum width=1cm, text width=1.3cm, inner sep=0]
\tikzstyle{tlabel}=[label, font=\scriptsize]

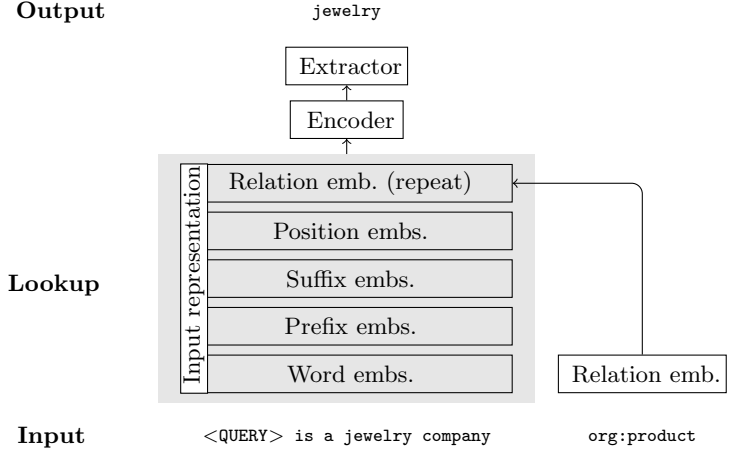
\begin{figure}
	\centering
	
	\begin{tikzpicture}
		\node (greybox) [fill=gray!20, minimum height=3.3cm, minimum width=5cm] at (0,0) {};
		\node (suffixembs) [embedding] at (greybox) {Suffix embs.};
		\node (prefixembs) [embedding, below=1.2mm of suffixembs] {Prefix embs.};
		\node (wordembs) [embedding, below=1.2mm of prefixembs] {Word embs.};
		\node (posembs) [embedding, above=1.2mm of suffixembs] {Position embs.};
		\node (relembs) [embedding, above=1.2mm of posembs] {Relation emb. (repeat)};
		\node (relemb) [embedding, minimum width=0cm, right=6mm of wordembs] {Relation emb.};
		\draw [arrow] (relemb) |- (relembs);

		\node (input1) [tlabel, below=2mm of greybox] {\texttt{$<$QUERY$>$ is a jewelry company}};
		\node (input2) [tlabel] at (input1 -| relemb) {\texttt{org:product}};

		\node (encoder) [layer, above=2mm of greybox] {Encoder};
		\node (extractor) [layer, above=2mm of encoder] {Extractor};
		
		\draw [arrow] (greybox) -- (encoder);
		\draw [arrow] (encoder) -- (extractor);

		\node (output) [above=2mm of extractor, tlabel] {\texttt{jewelry}};

		\node (inputlabel) [left=12mm of input1, elabel] {Input};
		\node [elabel] (lookuplabel) at (inputlabel |- suffixembs) {Lookup};
		\node [elabel] (outputlabel) at (inputlabel |- output) {Output};
		\node (o1) at (-1.85,3.2) {};

		\draw [fill=white] (o1 |- relembs.north) rectangle (wordembs.south west) node[pos=0.5, rotate=90, font=\small] {Input representation};
	\end{tikzpicture}
	
	\caption{Lookup layer and architecture overview.}
	\label{fig:overview}
\end{figure}

A \emph{model} consists of the lookup layer followed by an encoder layer, followed by a decoder layer. The remainder of this section provides a detailed discussion of layer variants.

\subsection{Lookup layer}
\label{sec:lookup}
In our problem formulation (argument extraction), a query entity and relation of interest are provided to the model,
and the missing argument has to be found. 
The model is therefore conditioned on the query, 
and it has knowledge of the query position.
We indicate the query position through wildcarding, where we replace
the query by a special token \verb=<QUERY>=, and additionally
we also use \emph{position embeddings} to indicate the distance 
of other tokens to the query position.
The relation in question is already provided at this stage 
to the model through the 
learned \emph{relation embeddings}. There is one 
embedding per relation.

Specifically, the lookup layer provides embeddings for five
types of information useful for answer extraction
that are concatenated for 
each position in the input context
(see Figure~\ref{fig:overview}).
For input position $i$, the input representation vector $\vec e_i$ is a concatenation of vectors:\footnote{Vectors are column vectors by default. Semicolons $[\cdots;\cdots]$ indicate vertical stacking along the column-axis, and commas $[\cdots,\cdots]$ indicate horizontal concatenation along the row-axis.} \begin{equation}
\vec e_i = [\vec e(w_i); \vec e(p_i); \vec e(s_i); \vec e(i-j); \vec e(r)]
\end{equation}

\begin{itemize}
 \item \textbf{Word embeddings} (embedding size $100$). Words contained in the pretrained GloVe vectors\footnote{\url{https://nlp.stanford.edu/projects/glove/}} are initialized with those vectors, otherwise they are initialized randomly. The vector $\vec e(w_i)$ is the embedding of $w_i$, the word at position $i$.
 \item \textbf{Affix embeddings.} Prefix and suffix embeddings (length: $2$ characters, embedding size: $100$) are learned in order to capture simple part-of-speech or named entity type generalization patterns (capitalization, morphological indicators). The vectors $\vec e(p_i)$ and $\vec e(s_i)$ are the embeddings of the prefix and suffix of $w_i$.
 \item \textbf{Position embedding.}
Since the first experiments using convolutional neural networks (CNNs) for relation extraction \cite{collobert2011natural,dossantos2015classifying} encoding the relative position to relation arguments has been key to good performance. We encode the relative position with respect to the query. Position encoding is used for all extractors, not only CNNs. The vector $\vec e(i-j)$ is the embedding of the relative position ($i-j$) w.r.t. the query position ($j$). The position embedding has size $10$.
 \item The \textbf{relation embedding} identifies the relation to the model and is repeated for every position in the input context. The vector $\vec e(r)$ is the embedding of the relation $r$.  The relation embedding size is set to $12$, the number of relations.
\end{itemize}

We denote the dimensionality of the input representation as $k$ ($k=3*100+10+12=322$). All embedding vectors are fine-tuned during training. The $k \times n$ matrix containing the input representations for all $n$ positions is denoted by $E=[\vec e_1, \cdots, \vec e_n]$.
\subsection{Encoders}
The sentence encoder translates the output of the lookup layer with neural network architectures that consider a wider context. We use three different alternative instantiations.
\subsubsection{RNN encoder}
In the recurrent neural network (RNN) encoder architecture, each candidate sentence is encoded by two layers of bi-directional Gated Recurrent Units (GRU) \cite{chung2014empirical} with a hidden size of $200$ ($100$ per direction). The hidden representation for position $i$ in the first GRU layer is the concatenation of a left-to-right and a right-to-left GRU hidden state. It is denoted by:
\begin{equation}
\vec h^{(1)}_i = [ \overrightarrow{\vec h}^{(1)}_i; \overleftarrow{\vec h}^{(1)}_i]
\end{equation}

Where the GRU hidden states are computed via the recurrences:
\begin{equation}
\overrightarrow{\vec h}^{(1)}_i = GRU(\overrightarrow{\vec h}^{(1)}_{i-1}, \vec e_i)
\end{equation} 
\begin{equation}
\overleftarrow{\vec h}^{(1)}_i = GRU(\overleftarrow{\vec h}^{(1)}_{i+1}, \vec e_i)
\end{equation} 

The second layer GRU takes the first layer as input and computes $\vec h^{(2)}_i~=~[ \overrightarrow{\vec h}^{(2)}_i; \overleftarrow{\vec h}^{(2)}_i]$ accordingly:
\begin{equation}
\overrightarrow{\vec h}^{(2)}_i = GRU(\overrightarrow{\vec h}^{(2)}_{i-1}, \vec h^{(1)}_i)
\end{equation} 
\begin{equation}\overleftarrow{\vec h}^{(2)}_i = GRU(\overleftarrow{\vec h}^{(2)}_{i+1}, \vec h^{(1)}_i)
\end{equation}

We did not observe a significant increase in performance on development data when using more layers, so the encoder output for the RNN encoder is $\vec h^{RNN}_i=\vec h^{(2)}_i$.
\subsubsection{CNN encoder}
CNNs are used with padding such that the number of input steps equals the number of output steps. We use 4 different filter widths: 3, 5, 7 and 9. For each filter width, we stack 3 layers with 32, 64 and 128 filters respectively. The ReLU activation is applied to each filter, and dropout (drop probability of $0.2$) is applied between the convolutional layers. The outputs of the last layer (for each filter width), and the relation embedding, are concatenated and used as input for the answer extractor.

More specifically, for filter width 3, the first layer CNN computes a 32-dimensional representation vector $\vec h^{(1;3)}_i$ (we write $\vec h^{(1;x)}_i$ for filter width x)
where each entry~$\vec h^{(1;3)}_{i,f}$ is computed from the input representation 
using the $3~*~k$~-~dimensional weight vector $\vec w^{(1;3)}_f$ for a particular filter $f$, and the ReLU activation:\footnote{We omit the bias term in affine transformations for readability.}
\begin{equation}
\vec h^{(1;3)}_{i,f} = ReLU(\vec w^{(1;3)T}_f \vec e_{[i-1:i+1]})
\end{equation}

where $\vec e_{[i-1:i+1]} = [\vec e_{i-1}; \vec e_{i}; \vec e_{i+1} ]$ and ReLU is defined component-wise as $ReLU(\vec x) = max(\vec 0,\vec x)$.

The second (and third) layer CNN computes a representation of size 64 (and 128) using the analogous formula:
\begin{equation}
\vec h^{(2;3)}_{i,f} = ReLU(\vec w^{(2;3)T}_f \vec h^{(1;3)}_{[i-1:i+1]})
\end{equation}

(Respectively $\vec h^{(3;3)}_{i,f} = max(0, \vec w^{(3;3)T}_f \vec h^{(2;3)}_{[i-1:i+1]})$ for the final third layer.)

The analogous formulas are applied for filter widths 5,7 and 9 (only considering wider contexts $[i-2:i+2]$ \emph{etc}). The final output of the CNN encoder is the concatenation of the 3rd layer output for each filter width. For the CNN architecture (but not for the other encoders), we observed small improvements on the development data by again concatenating the relation embeddings at each position:
\begin{equation}
\vec h^{CNN}_i = [\vec h^{(3;3)}_i; \vec h^{(3;5)}_i; \vec h^{(3;7)}_i; \vec h^{(3;9)}_i; \vec e(r)]
\end{equation}
\subsubsection{Self-attention encoder}
A third encoder uses the multi-headed self-attention architecture of \citeA{vaswani2017attention} to get an encoding for each position in the sequence. In self attention, the input representation for each position is used as a query to compute attention scores for all positions in the sequence. Those scores are then used to compute the weighted average of the input representations.

In multi-headed self-attention, input representations are first linearly mapped to lower-dimensional spaces, and the output vectors of several attention mechanisms (called \emph{heads}) are concatenated and form the output of one multi-headed self-attention layer. An attention head $a$ encodes a sequence of input vectors into a sequence of output vectors $\vec h_i^{(a)}$. Different heads pay attention to (i.e., put weight on) different parts or interactions in the input sequence. Different heads are parametrized independently (the respective parameters are marked by a superscript~$(a)$ to indicate that they are head-specific).

For one attention head $a$ in the first self-attention layer, we obtain the vector for position $i$:
\begin{equation}
\vec h^{(a)}_i = \Attention(W^{q(a)} \vec e_i, W^{K(a)} E, W^{V(a)} E)
\end{equation}
where $W^{q(a)}, W^{K(a)} , W^{V(a)}$ are linear transformations (matrices specific to head~$a$) to map the input representation into lower-dimensional space, and the matrix $E = \left[ \vec e_1, \dots, \vec e_n \right]$ is the matrix that contains the input representation (e.g., from the lookup layer, Section \ref{sec:lookup}). 
The function computing the resulting vector (from $\vec q~=~W^{q(a)} \vec e_i$, $K= W^{K(a)} E$ and $V=W^{V(a)} E$) is defined by:
\begin{equation}
\Attention(\vec q,K,V) = V \softmax(K^T \vec q)
\end{equation}

\begin{figure}
\includegraphics[width=0.75\textwidth]{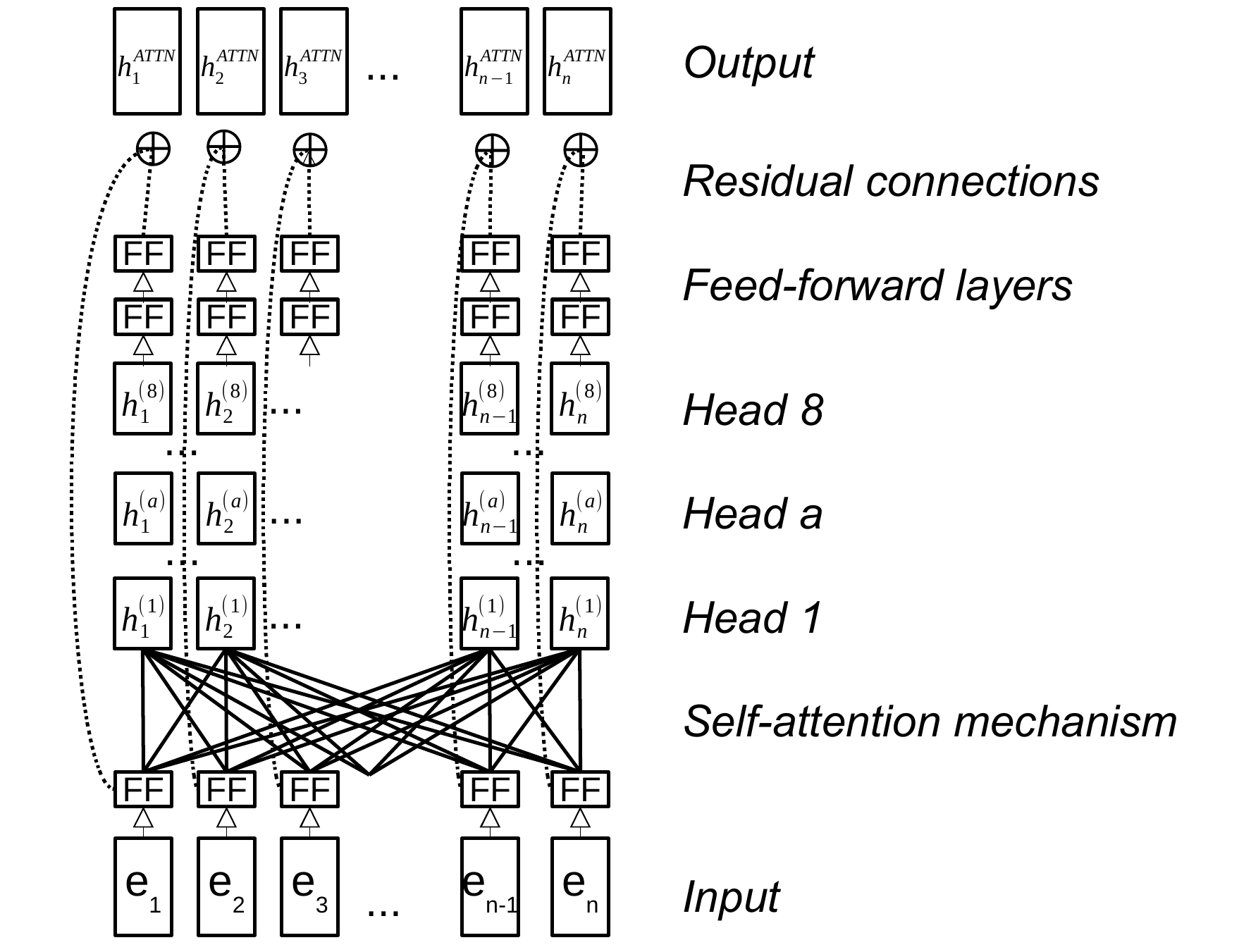}
	\caption{Schematic diagram of self-attention layer.}
	\label{fig:self_attention}
\end{figure}

We follow the setup described in \citeA{vaswani2017attention} and use 8 attention heads (each with a hidden size of $25$ resulting in an overall hidden size of $200$). 
The input to the self-attention mechanism is transformed by a feed-forward layer (output size 200, ReLU activation), and the output of the attention heads at each position is followed by two feed-forward layers (output sizes 400 and 200, ReLU activations)
One self-attention layer (the combination of self-attention heads and feed-forward layers) is stacked 3 times. More repetitions did not yield significant improvements on development data. See figure \ref{fig:self_attention} for a diagram depicting the architecture of one self-attention layer.

We deviated from the setup described in \citeA{vaswani2017attention} in the following ways, each of which improved the performance on the development data:

\begin{enumerate}
 \item We included residual connections that add the input of the self-attention mechanism directly to the output, rather than having two residual connections within each layer.
 \item We used batch normalization \cite{ioffe2015batch} rather than layer normalization \cite{ba2016layer} after concatenation of the attention heads and the MLP, respectively.
\end{enumerate}

As for the RNN and CNN encoders, the result is a vector representation $\vec h^{ATTN}_i$ for each position in the sentence.

\subsection{Extractors}
\label{subsec:extractors}

Extractors take the encoder output and predict the argument span (conditioned on the query entity and the relation of interest). If there is no argument for the relation of interest, the empty span is returned. We use three different architectures for argument extraction. In the following, the encoder output at position $i$ is denoted by~$\vec h_i$, irrespective of whether it stems from the RNN, CNN or self-attention encoder. The matrix $H$ represents the encoder outputs for all positions in the sentence, its dimensionality is length of the sentence times encoder output size.

\subsubsection{Pointer network}\label{ptr_network}

Pointer networks \cite{DBLP:conf/nips/VinyalsFJ15} are a
simple method to point to positions in a sequence by
calculating scores (similar to attention), normalizing them
using softmax and taking the argmax. In our case, 
two pointers are predicted, pointing to the
start and end positions of the
relation argument.

Figure \ref{fig:extractor} (left third) shows the processing flow for the
pointer network.
First, a \emph{summary vector} $\vec s$ is computed for the whole sentence
by max-pooling over the sentence encoder representation
(output of ``Encoder'' in
Figure \ref{fig:overview}), and
applying a fully connected layer with \verb=ReLU=
activation:
\begin{equation}
\bar{\vec s} = ReLU(W^s Pool(H))
\end{equation}
where $Pool$ returns a vector containing the row-wise
maximum of a matrix and $W^s$ is a
learned affine transformation.

A binary \textbf{label} is predicted through logistic
regression from the summary vector $\bar{\vec s}$; 
this label indicates whether the sentence contains an answer argument or not. 
The summary vector $\bar{\vec s}$ is also used as a context vector to compute
the pointer scores for predicting the \textbf{start
  position}, in a way similar to  attention modeling. For each
position in the sentence, the summary vector is
concatenated with the encoder output representation $\vec h_i$ at this
position, 
and from this a score is predicted (using a MLP with one hidden layer of size
$200$)
indicating how strongly this position should be associated with the start
of a relevant argument.
The softmax gives a distribution over the start positions:
\begin{equation}
p(start=i) = \softmax(MLP([\bar{\vec s}; \vec h_i]))
\end{equation}


The \textbf{end position} is predicted by the same
mechanism, but in this case the context vector is not the
summary vector $\bar{\vec s}$. 
Instead, the softmax
distribution over start positions output by the previous step
is used as the context vector (and concatenated with the encoder
outputs $\vec h_i$ for score prediction).
For sentences that do not contain an answer argument,
the start
and end positions are set to point to the query entity position during training.
This way
we hope to bias predictions to be closer to the query entity
position. At test time we exclude any predictions where
either the probability that an answer is less than or equal to 0.5,
or where the span overlaps with the query entity position.

\tikzstyle{square} = [rectangle, minimum width=6mm, minimum height=6mm, draw, inner sep=0]
\tikzstyle{vec} = [square, minimum height=9mm]
\tikzstyle{circle-label} = [circle,  draw, minimum size=4mm, inner sep=1pt]

\begin{figure*}
		\centering
			\begin{adjustbox}{width=1.0\textwidth}
			\begin{tikzpicture}
		
			\node at (0, 0) (pict2) {
				\begin{tikzpicture}
	
				\node[vec] (sq1) at (0, 0) {\small $\vec h_1$};
				\foreach \i in {2, ..., 6} {
				\pgfmathparse {int(\i-1)};
				\node[vec] (sq\i) [right=0mm of sq\pgfmathresult] {\small $\vec h_\i$};
				}
				\foreach \i in {1, 2, 6} {
			
				\node[circle-label] (label\i) [above=5mm of sq\i] {\small O};
				\draw[arrow] (sq\i) -- (label\i);
			}
						\foreach \i in {3, 4, 5} {
			\node[circle-label] (label\i) [above=5mm of sq\i, fill=gray!40] {\small \textbf{I}};
							\draw[arrow] (sq\i) -- (label\i);
		}
	\foreach \i in {1,..., 5} {
			\pgfmathparse {int(\i+1)};
	\draw (label\i) -- (label\pgfmathresult);
} 
				\end{tikzpicture}
				
			};
		
	\node [left=0.5cm of pict2] (pict1) {
	\begin{tikzpicture}
			
			\node[vec, draw = none] (sq-0) at (0, 0) {};
			\foreach \i in {1,...,6}
				{
					\pgfmathparse{int(\i-1)}
					\node [vec, right=0mm of sq-\pgfmathresult] (sq-\i) {\small $\vec h_\i$};
				}
			\node[vec, draw = none, right=0mm of sq-6] (sq-7) {};

				\node [above=2mm of sq-2] (pooling) {Pooling};
				\node [vec, above=12mm of sq-1.north east] (summary) {\small $\vec s$};

				\node [right=6mm of summary, align=center] (label) {Contains Answer:\\ Y/N};
				\draw [arrow] (summary) -- (label);
				\draw (sq-1.north west) -- (summary.south west);
				\draw (sq-6.north east) -- (summary.south east);

				\node (p0) at (0,3.2) {};
				\node (p1) [above=2pt of p0] {};
				\node (p2) [above=4pt of p0] {};
				\node (p6) [above=12pt of p0] {};
				\node (p4) [above=8pt of p0] {};

			

				\draw [dotted] (p0 -| sq-0.east) -- (p0 -| sq-7.center);
				\draw [dotted] (p0 -| sq-0.east) -- (p6 -| sq-0.east);
				
				\node (q0) at (0,4.7) {};
				\node (q1) [above=2pt of q0] {};
				\node (q2) [above=4pt of q0] {};
				\node (q6) [above=12pt of q0] {};
				\node (q4) [above=8pt of q0] {};
				\draw [fill=gray!40, draw=none] (p0 -| sq-3.west) rectangle (p6 -| sq-3.east);
				\draw [fill=gray!40, draw=none] (q0 -| sq-5.west) rectangle (q6 -| sq-5.east);

				\draw (p1 -| sq-1.west) -- (p1 -| sq-1.east) -- (p2 -| sq-2.west) -- (p2 -| sq-2.east) -- (p6 -| sq-3.west) -- (p6 -| sq-3.east) -- (p4 -| sq-4.west) -- (p4 -| sq-4.east) -- (p0 -| sq-5.west) -- (p0 -| sq-5.east) -- (p2 -| sq-6.west) -- (p2 -| sq-6.east) -- (p0 -| sq-7.west);
				\node (pdf1) [rotate=90, left=1mm of p6] {\small p(t)};
				\node (pdf2) [rotate=90, left=1mm of q6] {\small p(t)};
				\node (pdf3) [below=1mm of p0 -| sq-4.west] {\small t};
				\node (pdf4) [below=1mm of q0 -| sq-4.west] {\small t};
				\node (pdf5) [below=3mm of pdf1 -| sq-7.west] {start};
				\node (pdf6) [below=3mm of pdf2 -| sq-7.west] {end};


				
				\draw (q2 -| sq-1.west) -- (q2 -| sq-1.east) -- (q1 -| sq-2.west) -- (q1 -| sq-2.east) -- (q0 -| sq-3.west) -- (q0 -| sq-3.east) -- (q2 -| sq-4.west) -- (q2 -| sq-4.east) -- (q6 -| sq-5.west) -- (q6 -| sq-5.east) -- (q4 -| sq-6.west) -- (q4 -| sq-6.east) -- (q0 -| sq-7.west);
				\draw [dotted] (q0 -| sq-0.east) -- (q0 -| sq-7.center);
				\draw [dotted] (q0 -| sq-0.east) -- (q6 -| sq-0.east);
				\draw [arrow] (summary.north) -- (pdf3 -| summary);
				\draw [arrow] (p6.north -| summary) -- (pdf4 -| summary);

			\end{tikzpicture}
};

\node [right=1cm of pict2] (pict3) {
		\begin{tikzpicture}

		\node[square, draw=none] (sq-00) at (0, 0) {};
		
		
		\foreach \i in {1, ..., 6} {
			\pgfmathparse {int(\i-1)};
			\node[square] (sq-0\i) [right=0mm of sq-0\pgfmathresult] {\small $\vec h_\i$};
			\node[square] (sq-\i0) [below=0mm of sq-\pgfmathresult0] {\small $\vec h_\i$};
		}
	
			\foreach \i in {1, ..., 6} {
				
				\foreach \j in  {\i, ..., 6} {
					\pgfmathparse {int(\i-1)};
					\node[square, draw=none] (sq-\i\j) [below=0mm of sq-\pgfmathresult\j] {\small O};
				}
			}
		\node[square, draw=none, fill=gray!40, below=0mm of sq-25] {\small \textbf{I}};
	
		\draw (sq-06.south east) -- (sq-66.south east);
		
		\foreach \i in {1, ..., 6} {
			\draw (sq-\i\i.south west) -- (sq-\i\i.south east);
		}
	\foreach \i in {2, ..., 6} {
		\draw (sq-\i\i.north west) -- (sq-\i\i.south west);
	}

   \node[above=0.03cm of sq-03.north east] {end index};
   \node[above=0.1cm of sq-30.south west, rotate=90] {start index};	
   
		\end{tikzpicture}
	};

\node [above=3mm of pict1] (lab1) {Pointer Network};
\node (lab2) at (lab1 -| pict2) {Neural CRF Tagger};
\node (lab3) at (lab1 -| pict3)  {Table Filling};

			\end{tikzpicture}
		\end{adjustbox}

			\caption{Three extractor frameworks described in Section \ref{subsec:extractors} for predicting an answer span (from position 3 to 5 in this example, indicated by gray shading) from the encoder outputs $\vec h_i$.}
		\label{fig:extractor}
	\end{figure*}
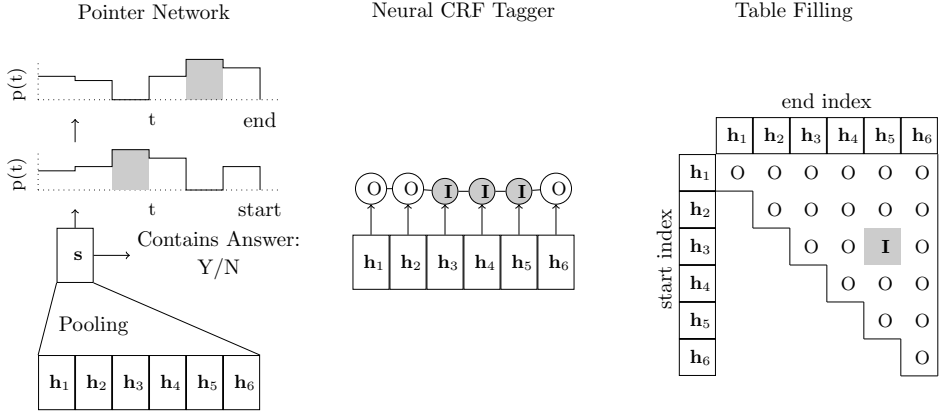

\subsubsection{CRF tagger}

The Conditional Random Field (CRF) tagger model predicts the answer span by predicting the label \verb="I"= for the answer, and \verb="O"= otherwise. As in previous work combining neural networks with CRFs \cite{collobert2011natural,lample2016neural}, the CRF combines local label scores, obtained from the features of the previous layers, with learned transition weights in order to obtain sequential label consistency: For an entire label sequence $\mathbf y = (y_1, y_2, \cdots, y_n)$ the global score is defined as:
\begin{equation}
s(H,\vec y)= \sum_{i=0}^n A_{y_i, y_{i+1}} + \sum_{i=1}^n s_{i,y_i}
\end{equation}
where $A$ is a (learned) matrix of transition scores from label $y_i$ to label $y_{i+1}$ (a special start label $y_0$ is assumed), and $s_{i,y_i}$ is the local label score for label $y_i$, obtained by a (learned) linear mapping from $\vec h_i$.

Viterbi decoding is used to find the predicted answer spans. The local label scores are also used in our system to assign a confidence value and to find the most likely answer span if there are several predicted spans.

\subsubsection{Table filling}

The table filling extractor jointly looks at pairs of sentence positions, and decides for each pair whether they are start and end positions for the query and relation on which the network is conditioned.
For a start position $i$ and an end position $j$, the table filling model decides whether those positions describe the start and end of the sought answer.
The table filling model uses the encoder outputs $\vec h_i$, $\vec h_j$ as the input for this binary decision (\texttt{I}: subspan is answer, \texttt{O}: subspan is not an answer, see Figure \ref{fig:extractor}, right diagram).

Compared to the pointer network (three model outputs: label, start, end) and the CRF tagger (number of model outputs = length of sequence), the table filling model has the most number of outputs to predict, as it needs, in principle, to pair each position in a sentence with all other (subsequent) positions in the sentence. To reduce the amount of computation that follows from this quadratic complexity, we limit the maximum length of representable answers to be 5 (which covers $98\%$  of actually occurring answers).
Note that
-- even though we exclude a large number of ``negative''
cells from the table and do not do any prediction for them --
the vast majority of output cells still has the negative
label (all but 1 pair of positions is not a relevant
relation argument), introducing a strong bias which may make
it harder for the model to predict a positive label at
all. 
For the combination with the CNN encoder,
it was necessary to
double the weights for the positive class,
following \citeA{gulcehre-EtAl:2016:P16-1}, 
to
deal with the highly skewed distribution of 
output classes (otherwise 
the table filling model would predict no answers).



For each pairing $(i,j)$ of potential start and end positions, we
concatenate the encoder vectors for the two positions, and
predict the corresponding cell value of the table. 
Logistic regression is used for cell prediction :
\begin{equation}
p(is\_answer=True|start=i, end=j) = \sigma([\vec h_i; \vec h_j]^T \vec w^{(table)})
\end{equation}
where a different weight vector $\vec w^{(table)}$  is learned for each answer length.

We experimented with deeper architectures for
cell value prediction, but did not observe any improvements, presumably due to the overwhelming majority of cells with a negative label.

\subsection{Hyper-parameters}

The following hyper-parameters were tuned on the development data (according to instance level accuracy) \cite{bengio2012practical} over the ranges given below. For tuning, the encoders were paired with the pointer network extractor (which is most similar to the Bidirectional Attention-Flow baseline, Section \ref{sec:bidaf}). We did not tune any hyperparameters specific to the extractors.

\begin{itemize}
 \item learning rate: $\{ 0.1, 0.01, 0.001 \}$
 \item number of CNN/GRU/Self Attention layers: $\{ 1, 2, 3, 4\}$
 \item CNN, maximal window size: $\{ 5, 7, 9 \}$
 \item CNN, maximal number of filters: $\{ 64, 128, 256 \}$
 \item Self-Attention, output size\footnote{Following \citeA{vaswani2017attention}, we use 8 heads.} (=number of heads * head size): $\{ 50, 100, 200, 400\}$
 \item GRU, hidden size: $\{ 50, 100, 200, 400 \}$
\end{itemize}

The resulting hyper-parameter choices are reported in the Sections describing the respective submodels. We use the 100-dimensional pretrained GloVe vectors of \citeA{pennington2014glove} and did not experiment with other word vector variants. The the size of the relation vector is equal to the number of relations (12, as for one-hot-encoding, but with the flexibility to arrange similar relations closer to each other in embedding space). We found that for the position embedding size a value equal to the square root of the maximum relative distance (in our experiments 10) gave good performance, and increasing it further did neither improve nor hurt the model.
All models use the Adam optimizer, the best value for learning rate was $0.01$ for all models. We found that larger batch sizes in general yielded better results than smaller ones, resulting in a batch size of 512 (which was the largest we could efficiently process on our infrastructure).

\begin{table*}
\begin{tabular}{p{3.6cm} p{9cm}}
\hline
{relation} & {example sentence} \\\hline

\verb=per:occupation=& $[$\emph{Alan Aubry}$]_{Q}$ ( born 24 September 1974 ) is a French $[$\emph{photographer}$]_{A}$ .\\
\verb=per:position_held=& Under pressure , former $[$\emph{T\'{a}naiste}$]_{A}$ $[$\emph{Erskine H. Childers}$]_{Q}$ agreed to run .\\
\verb=per:conflict=&It is named for $[$\emph{Henry Knox}$]_{Q}$ , an $[$\emph{American Revolutionary War}$]_{A}$ general .\\
\verb=per:notable_work=& In the $[$\emph{Steve Jackson}$]_{Q}$ Games card game $[$\emph{Munchkin}$]_{A}$ , there is a card called `` Dwarf Tossing '' .\\
\verb=per:participant_of=& $[$\emph{Ahlm}$]_{Q}$ was listed among the top ten goalscorers at the $[$\emph{2008 Olympics}$]_{A}$ tournament .\\
\verb=per:award_received=&$[$\emph{Alex Smith}$]_{Q}$'s name was put on the $[$\emph{Stanley Cup}$]_{A}$ in 1927 with Ottawa .\\
\verb=per:field_of_work=& While teaching at Berkeley , $[$\emph{John Harsanyi}$]_{Q}$ did extensive research in $[$\emph{game theory}$]_{A}$ .\\
  \verb=org:industry=& Select stores offer $[$\emph{fast food}$]_{A}$ outlets such as $[$\emph{Subway}$]_{Q}$ and Taco Bell .\\
\verb=per:noble_family=& Stefan was the son of Lazar and his wife $[$\emph{Milica}$]_{Q}$ , a lateral line of $[$\emph{Nemanjić}$]_{A}$ .\\
\verb=per:ethnic_group=&$[$\emph{Hamdi Ulukaya}$]_{Q}$ was born in 1972 to a $[$\emph{Kurdish}$]_{A}$ family in Turkey .\\
\verb=org:product=& $[$\emph{Lagos}$]_{Q}$ is a privately held American $[$\emph{jewelry}$]_{A}$ company\\
\verb=gpe:office =& Brown was the de facto $[$\emph{premier}$]_{A}$ of $[$\emph{Province of Canada}$]_{Q}$ in 1858 .
\\\hline
\end{tabular}
\caption{The table gives, for each relation, its name and
  an example sentence. (Query entity and correct answer entity are indicated in brackets.)}
\label{tab:data_samples}
\end{table*}

\begin{table}[!t]
{\center
\begin{tabular}{lrrrr}
\hline
 & train & dev & test \\\hline
\#{}instances& 673.677 & 340.050 &  335.883 \\
\#{}positive instances& 224.559 & 113.350 &  111.961 \\
\#{}fact triples & 132.983 & 66.925 & 66.697\\
\#{}query entities & 89.349 & 46.967 & 47.155
\\\hline
\end{tabular}
\caption{Number of instances, positive instances, distinct fact triples and distinct query entities for training development and test data.}
\label{tab:data_size}
}
\end{table}

\begin{table}[!t]
{\center
\begin{tabular}{llr}
\hline
relation & id & \#{}sentences \\\hline
\verb=per:occupation=& P106& 57693 \\
\verb=per:position_held=& P39& 47386\\
\verb=per:conflict=& P607& 20575\\
\verb=per:notable_work=& P800& 18826 \\
\verb=per:participant_of=& P1344& 14646\\
\verb=per:award_received=& P166& 13330\\
\verb=per:field_of_work=& P101& 13059\\
  \verb=org:industry=& P452& 12352\\
\verb=per:noble_family=& P53& 9260\\
\verb=per:ethnic_group=& P172& 7169\\
\verb=org:product=& P1056& 6482\\
\verb=gpe:office =& P1313& 3781
\\\hline
\end{tabular}
\caption{The table gives for each relation its name,
  Wikidata id and  number of training
  instances.}
\label{tab:data_samples_stats}
}
\end{table}

\section{Data set}
The models for predicting knowledge graph relations between entities that have non-standard type, proposed in the previous sections, are evaluated using a distantly supervised data set that we extracted from WikiData and Wikipedia specifically for this purpose.

We first identify relations  that meet three specific
criteria  and retrieve entity pairs 
for these relations. The three criteria 
are the following:

\begin{enumerate}[label=(\alph*)]
\item \textbf{Non-standard type.}
We look for relations that have one argument of a standard
type, the \emph{query}, and one argument of a non-standard
type, the \emph{slot}.
Training and evaluation are done for the task of identifying
correct fillers for the slot. 
We consider the MUC-7 named entity types (location, person, organization, money, percent, date, time) as standard types \cite{chinchor1997muc}.

 \item \textbf{Open class.} There must be a wide range of
   admissible values for the slot in question (i.e., the
   answers must be \emph{relational}, not \emph{categorical}
   \cite{hewlett2016wikireading}). For example, the
   WikiData relation P21 (\emph{sex or gender}) has a
   non-standard argument slot, but only a handful of
   distinct possible values are attested in WikiData; so P21
   is not a relation that we consider for our dataset. As a threshold, we
   require Wikidata to contain at least $1000$ distinct values for the slot in question.
 \item \textbf{Substantial coverage.} There must be a
   large number of facts (argument pairs) in Wikidata
   for a relation to be eligible for inclusion in our dataset. We require the minimal number of facts
   to be 10,000 for each relation.
\end{enumerate}

We check criterion (a) using the WikiData relation descriptions. We use the WikiData query interface\footnote{\url{https://query.wikidata.org/}} and the SPARQL query language to check  criteria (b) and (c), and to retrieve entity pairs (and their surface forms) for all relations. The relation entity pair tuples are randomly split into training ($50\%$), development ($25\%$) and test data ($25\%$).

In a second step, we retrieve sentences containing argument pairs
(\emph{distant supervision sentences}).
An English
Wikipedia dump (2016-09-20) is indexed on the sentence level using
ElasticSearch.\footnote{\url{https://www.elastic.co/}} For
each relation argument, aliases are obtained using Wikipedia
anchor text and the query expansion mechanism of the
RelationFactory KBP system
\cite{roth2014relationfactory}. Up to ten sentences are
retrieved for each argument pair. Although
criterion (c) 
requires a minimum number of 10,000 facts, we are not able
to find
a distant
supervision sentence for every pair. Therefore,
the number of actually occurring facts is less
than 10,000 for some relations. For each positive instance (sentence-relation-argument tuple), we sample two negative instances by replacing the relation with different relations (uniformly chosen at random).

Table \ref{tab:data_size} gives an overview of the training, development and test data sizes.
Table \ref{tab:data_samples_stats} lists the relations, together
with the number of training sentences for each relation. We
renamed the Wikidata ids to be more readable
 (similar to TAC KBP relations\footnote{\url{https://tac.nist.gov/about/index.html}}):
the names contain the entity type of the query argument as a
prefix.\footnote{The dataset and code are released at \url{http://cistern.cis.lmu.de/orae/}.}


\section{Experiments}
\subsection{Evaluation setup}
Each encoder architecture is combined with every extractor architecture. We compute accuracy, precision, and recall by assessing exact string match. We compute accuracy on a per-instance (query-relation-context) level. For precision, recall, and f-measure, two variants are computed: instance-level and tuple-level.

In the instance-level setup, the items which are considered are combinations of \emph{query-relation-context-answer} (where \emph{context} is a particular sentence represented by its id, and \emph{answer} is the missing argument that is to be extracted). In the tuple-level evaluation, the sentence id is ignored, and the same fact-tuple is counted only once, even if it has been extracted from several sentences, i.e., the items to be considered are combinations of \emph{query-relation-answer}. The tuple-level evaluation measures how well the ground-truth \emph{facts} are recovered, i.e., it corresponds to the quality of a knowledge graph obtained with the extraction algorithm, since repeated extractions are only counted once.

Precision and recall are computed from the sets of items, where $\mbox{relevant} = \mbox{set(correct items)}$ and $\mbox{retrieved} = \mbox{set(predicted items)}$; the f-measure is computed as usual $f=2pr/(p+r)$.

$$p = \frac{|\mbox{relevant} \cap \mbox{retrieved}|}{|\mbox{retrieved}|}$$
$$r = \frac{|\mbox{relevant} \cap \mbox{retrieved}|}{|\mbox{relevant}|}$$

\begin{table*}[!t]
\begin{tabularx}{\textwidth}{>{\hsize=0.58\hsize}X>{\hsize=.38\hsize}X>{\hsize=.38\hsize}X>{\hsize=.38\hsize}X>{\hsize=.38\hsize}X>{\hsize=.38\hsize}X>{\hsize=.38\hsize}X>{\hsize=.38\hsize}X>{\hsize=.38\hsize}X>{\hsize=.38\hsize}X}
  \hline 
   & \multicolumn{3}{c}{Pointer Network} & \multicolumn{3}{c}{Neural CRF Tagger} & \multicolumn{3}{c}{Table Filling} \\
   & \multicolumn{1}{c}{P} & \multicolumn{1}{c}{R} & \multicolumn{1}{c}{F} & \multicolumn{1}{c}{P} & \multicolumn{1}{c}{R} & \multicolumn{1}{c}{F} & \multicolumn{1}{c}{P} & \multicolumn{1}{c}{R} & \multicolumn{1}{c}{F} \\
   \hline
  ATTN & 75.50 & 73.51& 74.49
 &  72.39 & \underline{76.41}& 74.35
 & \underline{78.11}& 73.78& \underline{75.89} \\
CNN  & 78.23& \textbf{\underline{80.62}}& \textbf{79.41} & \textbf{\underline{82.59}}& 76.84& \underline{79.61} & \textbf{78.47}& 79.76& 79.11  \\
  RNN  & \textbf{78.80}& 79.17& 78.99 & \underline{82.53}& \textbf{81.19}& \textbf{\underline{81.86}} & 77.92& \textbf{\underline{81.44}}& \textbf{79.64}\\
\hline
\end{tabularx}
\caption{Each cell contains P/R/F on tuple
  level. The best values for each encoder (i.e., per row)
  are \underline{underlined}, the best values for each
  extractor (i.e., per column) are marked in \textbf{bold}.}
\label{tab:results}
\end{table*}

\subsection{Baselines}
\subsubsection{Argument extraction using Bidirectional Attention Flow}
\label{sec:bidaf}
\citeA{levy2017zero} formulate relation extraction as a reading comprehension problem: for each relation, a set of natural language questions is written by humans, and answers are extracted using the Bi-Directional Attention Flow (BiDAF) network \cite{seo2016bidirectional}.
In one of their experiments (the \emph{``KB Relations''}
setting), they do not provide the full questions, but rather
give the relation as an un-analyzed atom (the question
corresponds to the relation as the only pseudo-word). This
setting is applicable to our problem definition (and is simultaneously their best performing setup), hence we choose this system as a baseline. 
Since \citeA{levy2017zero} adapted a question answering
model to the task of relation answer extraction, some parts
of the model setup that help with analyzing natural language
questions (such as the attention mechanism that aligns parts
of the sentence with parts of the question) are superfluous
and not helpful for our task. A number of elements of 
BiDAF are similar to our model, but instantiated in a
different way. (i)  \citeA{seo2016bidirectional} use
character embeddings, we use prefix and suffix
embeddings. 
(ii) In BiDAF, attention is driven by the
query.
In one of our settings, we use self-attention where
any input information (words or relation) can recombine
information from the whole sentence. (iii) Similar to the prediction of
start and end points in \citeA{seo2016bidirectional}, one of our architectures is a pointer
network. We compare this to two other design choices for
predicting the answer span.

\subsubsection{Relation classification using Positional Attention}
\label{sec:posatt}

We also compare to the Position-aware Attention (PosAtt) model of \citeA{zhang2017position}, a strong relation classifier that can be used in a pipelined setting.
The PosAtt model requires as input a sentence with the \emph{query} and an already identified (by sequence tagging or string matching) \emph{answer candidate}.
PosAtt encodes this input with a neural architecture that summarizes the sentence using an attention mechanism that is aware of query and answer candidate positions, and predicts a relation for the encoded sentence.

Since the relations in ORAE are of non-standard type, and cannot be detected by off-the-shelf named entity taggers, we identify answer candidates by string matching:
Potential answers for a relation are all substrings in a sentence that were arguments for that relation in the training data.




\begin{table}[!t]
{\center
\begin{tabularx}{0.8\textwidth}{ >{\hsize=0.23\hsize}X >{\hsize=0.11\hsize}X >{\hsize=0.11\hsize}X >{\hsize=0.11\hsize}X >{\hsize=0.11\hsize}X  >{\hsize=0.11\hsize}X >{\hsize=0.11\hsize}X >{\hsize=0.11\hsize}X}
\hline
& \multicolumn{3}{c}{tuple level} & \multicolumn{4}{c}{instance level}  \\
& \multicolumn{1}{c}{P} & \multicolumn{1}{c}{R} & \multicolumn{1}{c}{F} & \multicolumn{1}{c}{P} & \multicolumn{1}{c}{R} & \multicolumn{1}{c}{F} & \multicolumn{1}{c}{Acc}\\
  \hline 
BiDAF & 70.86& 78.76& 74.60 & 76.35 & 75.84& 76.10& 90.11 \\
PosAtt & 83.65& 72.11& 77.45& -- & -- & -- & -- \\
CNN/CRF  & \textbf{82.59}& 76.84& 79.61 & \textbf{86.25}& 73.48& 79.35& 90.02\\
RNN/Table  &  77.92& \textbf{81.44}& 79.64 & 82.31& \textbf{78.38}& 80.30& 91.03 \\
RNN/CRF & 82.53& 81.19& \textbf{81.86} & 86.20& 78.25& \textbf{82.03}& \textbf{91.55}\\
\hline
\end{tabularx}
\caption{Comparison with Levy et al.\ BiDAF model, and Zhang et al. PosAtt model applied to our task. Reported results are P/R/F on tuple level and P/R/F/Acc on instance level. Best results per column marked in \textbf{bold}.}
\label{tab:BiDAF}
}
\end{table}

\subsection{Results and analysis}

\subsubsection{Architecture comparison}
Table \ref{tab:results} compares all combinations of encoder and extractor architectures introduced in the previous sections. In order to keep the overview uncluttered, we only show tuple-level results in Table \ref{tab:results}. See Table \ref{tab:BiDAF} for additional instance-level results for selected architectures.

\textbf{Encoders.} For the encoder architectures, one can see that the self-attention mechanism (ATTN) is the weakest (although competitive to the baselines, see below), reaching an f-measure of $75.89$ in the best combination.\footnote{Unless indicated otherwise, we discuss tuple-level scores.}

Good results are obtained by the CNN encoder, with the f-measure reaching $79.61$ (and with similar results obtained when different extractors are chosen). A slightly higher f-measure of $81.86$ is achieved with the RNN encoder, however, for this encoder, results vary more depending on the choice of extractor.

Compared to RNN and CNN, \textbf{self-attention modeling} is the \emph{least local} of all three encoders, as it can incorporate information from the entire sentence by the same mechanism; positional information is only captured via the positional embeddings. The comparatively weak performance of the ATTN encoder indicates that some locality bias may be beneficial for argument extraction (higher influence of neighboring words, distance to query), and that non-local modeling, only relying on positional embeddings, is not sufficient.

The \textbf{CNN encoder} is the \emph{most local} of all encoders: information of neighboring words is combined using the stacked filters. The only long-range dependency that can be captured is the distance to the query (via positional embeddings). The relatively good results of the CNN encoder indicate that most relevant information can be captured by this mechanism.

The \textbf{RNN encoder} can use \emph{all non-local information} via its bidirectional recurrences, but at the same time RNNs have a bias towards local information as it needs to go through fewer transformations. In our experiments this way of encoding the entire sentence information via RNNs yields the best results overall.

\textbf{Extractors.} The \textbf{pointer network} is for none of the encoders the best extractor. However, differences to the other extractors are relatively small. The limitation of the pointer network is that decisions for start and end position are not optimized jointly (the score distribution over end positions cannot influence that over start positions), and this fact may limit the model to gain the last percentage points of extra performance needed.

The \textbf{neural CRF tagger} is the best extractor for both the CNN and the RNN encoder, achieving the best results overall. Start and end position are jointly modeled and globally optimized via the tag sequence and the transition scores.

The \textbf{table filling extractor} models start and end positions jointly by design. The biggest difficulty for the table filling extractor is the fact that the number of negative labels (combinations of start and end positions that do not constitute a correct answer) grows quadratically with the sentence length. Without correcting for this imbalance by doubly weighting positive labels in the objective function, recall values would be extremely low -- for the CNN encoder without this reweighting no answer would be extracted at all. Despite its relatively good performance, the table filling extractor is therefore less stable than the pointer network or CRF extractor.

\textbf{Lookup layer.} We include an ablation analysis, to examine how different input representations interact with encoder layers and end-to-end models. For each encoder architecture, we take its best combination with a decoder and compare its performance using the full input representation and its performance with a reduced input representation (in terms of tuple-level f-measure), we report this difference in Table \ref{tab:ablation}. We ablate word embeddings, affix embeddings, position embeddings, and we compare to a setup where the query is not wildcarded. We also compare to a setup 
where the relation of interests was not given to the model (i.e. the model loses the capability to distinguish between different relations).

The CNN and RNN models rely more on \textbf{word embeddings}, while the the self-attention model relies more on \textbf{affix embeddings}.
\textbf{Position embeddings} are crucial for the self-attention encoder \cite{vaswani2017attention}, 
in contrast, CNN and RNN model sequential order by design and do not depend on position embeddings.
\textbf{Query wildcarding} is the most important factor in representing the input.
Without query wildcarding, the model may be prone to overfit the queries seen during training, and moreover the information about what element in the sentence is the query is passed on to the model only via the relative position embeddings.
Not surprisingly, \textbf{relation embeddings} are essential to the performance of the models.

\begin{table*}[!thbp]
\centering
\begin{tabular}{llllll }
  \hline
  \textbf{Architecture} & \textbf{Word} & \textbf{Affix} & 
  \textbf{Position} & \textbf{Query} & \textbf{Relation} \\
  \hline
  ATTN+Table  & 0.16  & 1.73  & 3.54 & 4.89 & 50.76 \\
  CNN+CRF  & 2.63  & 0.16  & 0.16 & 2.87 & 74.06 \\
  RNN+CRF  & 2.73  & 0.05 & -0.29 & 3.71 & 79.99 \\
  \hline
\end{tabular}
\caption{Performance of full input representation minus performance of reduced input representation, for Self Attention+Table Filling, CNN+CRF Tagger, RNN+CRF Tagger. Ablated elements: Word embeddings, affix embeddings, position embeddings, query wildcarding and relation embeddings.}
\label{tab:ablation}
\end{table*}

\subsubsection{Baselines} 
Table \ref{tab:BiDAF} shows the
performance of the BiDAF architecture adapted to relation extraction
as in \citeA{levy2017zero}
upon training and testing on the open-type
relation argument extraction task.
We provide a full comparison (precision, recall, f-measure, accuracy; instance and tuple level) of this baseline to our best-performing (by tuple-level f-measure) encoder-extractor architectures (RNN/CRF, CNN/CRF and RNN/Table).

The number of instances considered in PosAtt differs from that in answer extraction models (BiDAF and our approaches), since for one query and relation (an instance in answer extraction) there can be many or no answer candidates. We therefore only consider tuple-level scores for comparison with PosAtt. 
PosAtt does not have the freedom to predict any substring as an answer since it depends on answer candidate identification as a preceding step in a pipeline.
It consequently has the lowest recall of all considered models.
The good precision of PosAtt indicates, however, that it is a very strong relation classification model.

As for uninformed baselines (like NER-based pipelined systems, that cannot detect non-standard types), always predicting the empty answer would yield an accuracy of $66.67\%$. For the f-measure there is no simple uninformed baseline, so the base score for the f-measure would be close to $0$. Hence, all models perform quite well on the task, extracting answers with accuracies of $\sim 90\%$.

Clearly, our best
performing Neural CRF Tagger has approximately +7\% absolute
better f-measure in both instance and tuple wise
evaluation. We attribute the improvements of most of
our encoder-extractor based models to the following design choices:

\begin{itemize}
\item We wildcard the query entity
  (\verb=<QUERY>= in Figure \ref{fig:overview}).
 This directs extractors to focus their search for the slot filler on the
 vicinity of the query.
Since most answers occur close to the query, introducing
this bias improves performance. Wildcarding also prevents overfitting since the model
  cannot learn from the specific lexical material of the query.
\item The combination of prefix and suffix embeddings
is advantageous because most of the information about
possible nonstandard entity types that is not already
captured by word embeddings is captured by these two affixes.
\item BiDAF devotes modeling capacity to \emph{bidirectional attention} (in order to detect relevant parts of a question), which is irrelevant in the relation scenario since the ``question'' is represented as exactly one token, i.e., the relation itself. 
\item CRF and Table-filling answer extraction can model start and end positions jointly, while BiDAF predicts them independently.
\end{itemize}

To summarize, our experiments indicate that for relation argument extraction, an RNN network with a tagging based answer extractor is superior to extractors based on table filling or based on the prediction of start and end positions (as often done by question-answering systems such as BiDAF).




\subsubsection{Discussion}
We have extended and redefined the problem of
slot filling to the task of 
open type relation argument extraction (ORAE).
The type of model we have proposed to address
ORAE
is not just a model that solves relation classification (or
slot filling); it also jointly solves the task of
finding the entities.

There are several
advantages to
this extension and redefinition of slot filling.

\begin{itemize}
 \item In ORAE, the model can use all available information
   in the sentence and optimize decision thresholds for the
   task at hand (i.e., filler identification), avoiding
   tagging errors that it cannot recover from.
 \item In ORAE, the model can be trained by distant supervision. As long as there are surface strings of entity pairs from a knowledge base, the model can be trained. The co-occurrence requirement for two entities during training also provides some disambiguation and filtering of spurious matches.
 \item Our definition of ORAE treats standard and
   non-standard named entity types in completely the same way.
This enables us to  detect  non-standard slot fillers
like \emph{job titles}, \emph{products} and \emph{industries} that approaches
based on named entity tagging have difficulties with.
\end{itemize}

One shortcoming of the setup we presented in this article
is that only one  answer is predicted per query
instance. Although the model architecture can easily be reformulated for
a more general setting, the problem lies with the sparse distant
supervision training data that only rarely contains matches
with multiple answers within a given
context. Given this lack of training data, it is not clear how
the parameters of such a more general model should best be estimated.

\section{Related work}

In opinion recognition, early work has focused on extracting opinion holders and opinion items with CRFs and integer linear programming \cite{choi2006joint}. See \cite{culotta2006integrating} and \cite{hoffmann2010learning} for other approaches to argument tagging using traditional feature-based CRFs. This line of research has recently been extended \cite{katiyar2016investigating} to a neural tagging scheme, where relations (and the distance to the related token) are predicted per token by a long short-term memory network (LSTM, \cite{hochreiter1997long}). 
This setting is quite different from ours since prediction is not conditioned on a query entity; 
apart from the different problem formulation, this also implies that the model cannot be trained with incomplete annotation via distant supervision \cite{mintz2009distant}, since training needs all labels to be present (not just those for the query Q). \citeA{zheng2017joint} use a tagging scheme similar to \cite{katiyar2016investigating} to annotate relation arguments in sentences. They do not condition on a query entity and need to downweight non-argument labels to overcome sparsity in the training data.

Similarly, table filling models have been developed to extract entities and relations, see \cite{miwa2014modeling} for the original feature-based formulation and \cite{miwa2016end} for an RNN-based extension of the model. In contrast to our work, this model requires fully annotated data (no distant supervision), and therefore has only been applied to relations with standard named entities (\emph{person, location, organization}), where the motivation for open-type argument extraction is less strong. Another extension \cite{gupta2016table} obtained improvements by relying on already identified named entity spans. We compare a variant of neural table filling that does not rely on any of these conditions with a range of alternative argument extraction methods.

Wikireading \cite{hewlett2016wikireading} is the task of
extracting infobox properties from Wikipedia articles about
a certain entity (similar
to \cite{hoffmann2010learning}). Some aspects of
Wikireading are easier than the problem we are dealing
with, for example, it is guaranteed that there is an answer
for every paragraph in the dataset, and the query entity is
guaranteed to be the topic of the article. Other aspects are
more difficult, for example, only 46\% of the answers in the
data set are contained as exact strings, the majority has to
be inferred. In contrast, we are concerned with the problem
of predicting whether relations hold between mentions as
they are expressed in text.

Another approach to overcoming reliance on named entity
recognition in relation extraction is to do segmentation of text
heuristically based on part-of-speech patterns and
cooccurrences, and then to proceed in the traditional
instance-based paradigm \cite{ren2017cotype}.


Traditional relation classification and, more generally,
work deciding whether a relation holds between two
identified subparts of a sentence
is also relevant.
\citeA{collobert2011natural} combined CNNs with position
embeddings and CRFs for semantic role labeling.
Subsequent work confirmed that convolutional neural
networks are appropriate models for
relation classification
\cite{zeng2014relation,dossantos2015classifying,adel2016comparing,vu2016combining}. 
Other approaches have employed RNN variants for representing sentences for relation classification \cite{verga2015multilingual,xu2016improved}.

Another related field is that of question answering (QA).
The introduction of the Stanford Question Answering Dataset (SQuAD) \cite{rajpurkar-EtAl:2016:EMNLP2016}
has given rise to a large body of work on answer extraction.
\citeA{seo2016bidirectional} and \citeA{chen2017reading}
introduce an efficient method of aligning question and
paragraph words through an attention mechanism \cite{DBLP:journals/corr/BahdanauCB14} to obtain an answer span.
\citeA{wang2017rnet} propose
an architecture that, based on match LSTM,
builds a question aware passage representation and uses an attention-based pointer
network \cite{DBLP:conf/nips/VinyalsFJ15} to predict the start
and end positions of the answer.

Recently, \citeA{levy2017zero} presented
an approach that bridges question answering and
query-driven answer extraction.
They convert the traditional
entity-driven relation extraction to a QA setup by
crowd-sourcing knowledge base  relations into natural language
questions. They
utilize the bidirectional
attention flow networks (BiDAF) of \cite{seo2016bidirectional} to extract answers.
We compare our experimental results to this strong
baseline.
\section{Conclusion}
We have defined the task of
Open-Type Relation Argument Extraction (ORAE), where the
model has to extract relation arguments without being able
to rely on an entity extractor to find the argument
candidates. ORAE can be viewed as a type of
entity-driven slot-filling,  the task of identifying and
gathering relational information about a query entity
from a large corpus of text.
However, the most common approaches to slot-filling are
pipelined architectures, in which relation classification is an
isolated step that heavily relies on pre-processing modules
such as named entity recognition, to which a large part of
end-to-end errors can be attributed. Our approach to ORAE
has two conceptual advantages. First, it is
more general than 
slot-filling as it is also applicable to
non-standard named entity types that could not be dealt with
previously.
Second, while the problem we define
is more difficult than standard slot filling,
we eliminate 
an important source of errors: tagging errors that propagate
throughout the pipeline and that are notoriously hard to
correct downstream.

We have presented a distantly supervised data set for
training and evaluating ORAE models, based on WikiData
relations; the arguments in our dataset are non-standard type named
entities, e.g., notable work (which can be any title of a
book or other work of art) or
product (which can be any product name).

We have experimented with a wide range of neural network
architectures to solve ORAE, each consisting of
a \emph{sentence encoder}, which
computes a vector representation for every sentence
position, and an \emph{argument extractor}, which
extracts the relation argument from that representation.
We experimented with
convolutional neural
networks, recurrent neural networks, and self-attention
as sentence encoders; and
with
pointer network, conditional random fields tagging  and
table filling as argument extractors. Every encoder was combined with every
extractor, and high accuracy was obtained for most combinations.
The combination of recurrent neural network encoder with conditional random field extractor gave the best results, 
$+4\%$ absolute f-measure better than a state-of-the-art pipelined model based on argument matching,
and $+7\%$ absolute f-measure better than a previously proposed adaptation of a question answering model.

\bibliographystyle{chicago-nle}
\bibliography{bibliography_nle}

\begin{thebibliography}{}

\bibitem[\protect\citeauthoryear{Adel, Roth, and Sch{\"{u}}tze}{Adel
  et~al.}{2016}]{adel2016comparing}
Adel, H., B.~Roth, and H.~Sch{\"{u}}tze 2016.
\newblock Comparing convolutional neural networks to traditional models for
  slot filling.
\newblock In {\em {NAACL} {HLT} 2016, The 2016 Conference of the North American
  Chapter of the Association for Computational Linguistics: Human Language
  Technologies}, San Diego, California, USA, pp.\  828--38.

\bibitem[\protect\citeauthoryear{Angeli, Tibshirani, Wu, and Manning}{Angeli
  et~al.}{2014}]{angeli2014combining}
Angeli, G., J.~Tibshirani, J.~Wu, and C.~D. Manning 2014.
\newblock Combining distant and partial supervision for relation extraction.
\newblock In {\em Proceedings of the 2014 Conference on Empirical Methods in
  Natural Language Processing (EMNLP)}, Doha, Qatar, pp.\  1556--67.

\bibitem[\protect\citeauthoryear{Ba, Kiros, and Hinton}{Ba
  et~al.}{2016}]{ba2016layer}
Ba, L.~J., R.~Kiros, and G.~E. Hinton 2016.
\newblock Layer normalization.
\newblock {\em CoRR\/}~{\em abs/1607.06450}.

\bibitem[\protect\citeauthoryear{Bahdanau, Cho, and Bengio}{Bahdanau
  et~al.}{2014}]{DBLP:journals/corr/BahdanauCB14}
Bahdanau, D., K.~Cho, and Y.~Bengio 2014.
\newblock Neural machine translation by jointly learning to align and
  translate.
\newblock {\em CoRR\/}~{\em abs/1409.0473}.

\bibitem[\protect\citeauthoryear{Bengio}{Bengio}{2012}]{bengio2012practical}
Bengio, Y. 2012.
\newblock Practical recommendations for gradient-based training of deep
  architectures.
\newblock In {\em Neural networks: Tricks of the trade}, pp.\  437--478.
  Springer.

\bibitem[\protect\citeauthoryear{Chen, Fisch, Weston, and Bordes}{Chen
  et~al.}{2017}]{chen2017reading}
Chen, D., A.~Fisch, J.~Weston, and A.~Bordes 2017.
\newblock Reading {W}ikipedia to answer open-domain questions.
\newblock In {\em Proceedings of the 55th Annual Meeting of the Association for
  Computational Linguistics (Volume 1: Long Papers)}, Vancouver, Canada, pp.\
  1870--9.

\bibitem[\protect\citeauthoryear{Chinchor and Robinson}{Chinchor and
  Robinson}{1997}]{chinchor1997muc}
Chinchor, N. and P.~Robinson 1997.
\newblock Muc-7 named entity task definition.
\newblock In {\em Proceedings of the 7th Conference on Message Understanding
  \url{http://anthology.aclweb.org/M/M98/}}.

\bibitem[\protect\citeauthoryear{Choi, Breck, and Cardie}{Choi
  et~al.}{2006}]{choi2006joint}
Choi, Y., E.~Breck, and C.~Cardie 2006.
\newblock Joint extraction of entities and relations for opinion recognition.
\newblock In {\em Proceedings of the 2006 Conference on Empirical Methods in
  Natural Language Processing (EMNLP)}, Sydney, Australia, pp.\  431--9.

\bibitem[\protect\citeauthoryear{Chung, G{\"{u}}l{\c{c}}ehre, Cho, and
  Bengio}{Chung et~al.}{2014}]{chung2014empirical}
Chung, J., {\c{C}}.~G{\"{u}}l{\c{c}}ehre, K.~Cho, and Y.~Bengio 2014.
\newblock Empirical evaluation of gated recurrent neural networks on sequence
  modeling.
\newblock {\em CoRR\/}~{\em abs/1412.3555}.

\bibitem[\protect\citeauthoryear{Collobert, Weston, Bottou, Karlen,
  Kavukcuoglu, and Kuksa}{Collobert et~al.}{2011}]{collobert2011natural}
Collobert, R., J.~Weston, L.~Bottou, M.~Karlen, K.~Kavukcuoglu, and P.~P. Kuksa
  2011.
\newblock Natural language processing (almost) from scratch.
\newblock {\em Journal of Machine Learning Research\/}~{\em 12}, 2493--537.

\bibitem[\protect\citeauthoryear{Culotta, McCallum, and Betz}{Culotta
  et~al.}{2006}]{culotta2006integrating}
Culotta, A., A.~McCallum, and J.~Betz 2006.
\newblock Integrating probabilistic extraction models and data mining to
  discover relations and patterns in text.
\newblock In {\em {NAACL} {HLT} 2006, Proceedings of the Human Language
  Technology Conference of the North American Chapter of the Association of
  Computational Linguistics}, New York, USA, pp.\  296--303.

\bibitem[\protect\citeauthoryear{dos Santos, Xiang, and Zhou}{dos Santos
  et~al.}{2015}]{dossantos2015classifying}
dos Santos, C.~N., B.~Xiang, and B.~Zhou 2015.
\newblock Classifying relations by ranking with convolutional neural networks.
\newblock In {\em Proceedings of the 53rd Annual Meeting of the Association for
  Computational Linguistics and the 7th International Joint Conference on
  Natural Language Processing of the Asian Federation of Natural Language
  Processing}, Volume~1, Beijing, China, pp.\  626--34.

\bibitem[\protect\citeauthoryear{G{\"u}l{\c c}ehre, Ahn, Nallapati, Zhou, and
  Bengio}{G{\"u}l{\c c}ehre et~al.}{2016}]{gulcehre-EtAl:2016:P16-1}
G{\"u}l{\c c}ehre, {\c C}., S.~Ahn, R.~Nallapati, B.~Zhou, and Y.~Bengio 2016.
\newblock Pointing the unknown words.
\newblock In {\em Proceedings of the 54th Annual Meeting of the Association for
  Computational Linguistics}, Volume~1, Berlin, Germany, pp.\  140--9.

\bibitem[\protect\citeauthoryear{Gupta, Sch{\"u}tze, and Andrassy}{Gupta
  et~al.}{2016}]{gupta2016table}
Gupta, P., H.~Sch{\"u}tze, and B.~Andrassy 2016.
\newblock Table filling multi-task recurrent neural network for joint entity
  and relation extraction.
\newblock In {\em Proceedings of COLING 2016, the 26th International Conference
  on Computational Linguistics: Technical Papers}, Osaka, Japan, pp.\
  2537--47.

\bibitem[\protect\citeauthoryear{Hewlett, Lacoste, Jones, Polosukhin,
  Fandrianto, Han, Kelcey, and Berthelot}{Hewlett
  et~al.}{2016}]{hewlett2016wikireading}
Hewlett, D., A.~Lacoste, L.~Jones, I.~Polosukhin, A.~Fandrianto, J.~Han,
  M.~Kelcey, and D.~Berthelot 2016.
\newblock Wikireading: A novel large-scale language understanding task over
  {W}ikipedia.
\newblock In {\em Proceedings of the 54th Annual Meeting of the Association for
  Computational Linguistics}, Berlin, Germany, pp.\  1535--45.

\bibitem[\protect\citeauthoryear{Hochreiter and Schmidhuber}{Hochreiter and
  Schmidhuber}{1997}]{hochreiter1997long}
Hochreiter, S. and J.~Schmidhuber 1997.
\newblock Long short-term memory.
\newblock {\em Neural computation\/}~{\em 9\/}(8), 1735--80.

\bibitem[\protect\citeauthoryear{Hoffmann, Zhang, and Weld}{Hoffmann
  et~al.}{2010}]{hoffmann2010learning}
Hoffmann, R., C.~Zhang, and D.~S. Weld 2010.
\newblock Learning 5000 relational extractors.
\newblock In {\em Proceedings of the 48th Annual Meeting of the Association for
  Computational Linguistics}, Uppsala, Sweden, pp.\  286--95.

\bibitem[\protect\citeauthoryear{Ioffe and Szegedy}{Ioffe and
  Szegedy}{2015}]{ioffe2015batch}
Ioffe, S. and C.~Szegedy 2015.
\newblock Batch normalization: Accelerating deep network training by reducing
  internal covariate shift.
\newblock In {\em Proceedings of the 32nd International Conference on Machine
  Learning}, Volume~37, Lille, France, pp.\  448--56.

\bibitem[\protect\citeauthoryear{Katiyar and Cardie}{Katiyar and
  Cardie}{2016}]{katiyar2016investigating}
Katiyar, A. and C.~Cardie 2016.
\newblock Investigating {LSTMs} for joint extraction of opinion entities and
  relations.
\newblock In {\em Proceedings of the 54th Annual Meeting of the Association for
  Computational Linguistics}, Berlin, Germany, pp.\  919--29.

\bibitem[\protect\citeauthoryear{Lafferty, McCallum, and Pereira}{Lafferty
  et~al.}{2001}]{lafferty2001crf}
Lafferty, J.~D., A.~McCallum, and F.~C.~N. Pereira 2001.
\newblock Conditional random fields: Probabilistic models for segmenting and
  labeling sequence data.
\newblock In {\em Proceedings of the Eighteenth International Conference on
  Machine Learning {(ICML} 2001), Williams College, Williamstown, MA, USA, June
  28 - July 1, 2001}, pp.\  282--289.

\bibitem[\protect\citeauthoryear{Lample, Ballesteros, Subramanian, Kawakami,
  and Dyer}{Lample et~al.}{2016}]{lample2016neural}
Lample, G., M.~Ballesteros, S.~Subramanian, K.~Kawakami, and C.~Dyer 2016.
\newblock Neural architectures for named entity recognition.
\newblock In {\em {NAACL} {HLT} 2016, The 2016 Conference of the North American
  Chapter of the Association for Computational Linguistics: Human Language
  Technologies}, San Diego, California, USA, pp.\  260--70.

\bibitem[\protect\citeauthoryear{Levy, Seo, Choi, and Zettlemoyer}{Levy
  et~al.}{2017}]{levy2017zero}
Levy, O., M.~Seo, E.~Choi, and L.~Zettlemoyer 2017.
\newblock Zero-shot relation extraction via reading comprehension.
\newblock In {\em Proceedings of the 21st Conference on Computational Natural
  Language Learning (CoNLL 2017)}, Vancouver, Canada, pp.\  333--42.

\bibitem[\protect\citeauthoryear{Mintz, Bills, Snow, and Jurafsky}{Mintz
  et~al.}{2009}]{mintz2009distant}
Mintz, M., S.~Bills, R.~Snow, and D.~Jurafsky 2009.
\newblock Distant supervision for relation extraction without labeled data.
\newblock In {\em Proceedings of the Joint Conference of the 47th Annual
  Meeting of the ACL and the 4th International Joint Conference on Natural
  Language Processing of the AFNLP}, Volume~2, Suntec, Singapore, pp.\
  1003--11.

\bibitem[\protect\citeauthoryear{Miwa and Bansal}{Miwa and
  Bansal}{2016}]{miwa2016end}
Miwa, M. and M.~Bansal 2016.
\newblock End-to-end relation extraction using {LSTMs} on sequences and tree
  structures.
\newblock In {\em Proceedings of the 54th Annual Meeting of the Association for
  Computational Linguistics}, Berlin, Germany, pp.\  1105--16.

\bibitem[\protect\citeauthoryear{Miwa and Sasaki}{Miwa and
  Sasaki}{2014}]{miwa2014modeling}
Miwa, M. and Y.~Sasaki 2014.
\newblock Modeling joint entity and relation extraction with table
  representation.
\newblock In {\em Proceedings of the 2014 Conference on Empirical Methods in
  Natural Language Processing (EMNLP)}, Doha, Qatar, pp.\  1858--69.

\bibitem[\protect\citeauthoryear{Pennington, Socher, and Manning}{Pennington
  et~al.}{2014}]{pennington2014glove}
Pennington, J., R.~Socher, and C.~Manning 2014.
\newblock Glove: Global vectors for word representation.
\newblock In {\em Proceedings of the 2014 conference on empirical methods in
  natural language processing (EMNLP)}, pp.\  1532--1543.

\bibitem[\protect\citeauthoryear{Pink, Nothman, and Curran}{Pink
  et~al.}{2014}]{pink2014analysing}
Pink, G., J.~Nothman, and J.~R. Curran 2014.
\newblock Analysing recall loss in named entity slot filling.
\newblock In {\em Proceedings of the 2014 Conference on Empirical Methods in
  Natural Language Processing, {EMNLP} 2014}, Doha, Qatar, pp.\  820--30.

\bibitem[\protect\citeauthoryear{Rajpurkar, Zhang, Lopyrev, and
  Liang}{Rajpurkar et~al.}{2016}]{rajpurkar-EtAl:2016:EMNLP2016}
Rajpurkar, P., J.~Zhang, K.~Lopyrev, and P.~Liang 2016.
\newblock Squad: 100,000+ questions for machine comprehension of text.
\newblock In {\em Proceedings of the 2016 Conference on Empirical Methods in
  Natural Language Processing}, Austin, Texas, pp.\  2383--92.

\bibitem[\protect\citeauthoryear{Ren, Wu, He, Qu, Voss, Ji, Abdelzaher, and
  Han}{Ren et~al.}{2017}]{ren2017cotype}
Ren, X., Z.~Wu, W.~He, M.~Qu, C.~R. Voss, H.~Ji, T.~F. Abdelzaher, and J.~Han
  2017.
\newblock Cotype: Joint extraction of typed entities and relations with
  knowledge bases.
\newblock In {\em Proceedings of the 26th International Conference on World
  Wide Web}, Perth, Australia, pp.\  1015--24.

\bibitem[\protect\citeauthoryear{Roth}{Roth}{2015}]{roth2015effective}
Roth, B. 2015.
\newblock {\em Effective distant supervision for end-to-end knowledge base
  population systems}.
\newblock Ph.\ D. thesis, Saarland University.

\bibitem[\protect\citeauthoryear{Roth, Barth, Chrupala, Gropp, and Klakow}{Roth
  et~al.}{2014}]{roth2014relationfactory}
Roth, B., T.~Barth, G.~Chrupala, M.~Gropp, and D.~Klakow 2014.
\newblock {RelationFactory}: A fast, modular and effective system for knowledge
  base population.
\newblock In {\em Proceedings of the Demonstrations at the 14th Conference of
  the European Chapter of the Association for Computational Linguistics},
  Gothenburg, Sweden, pp.\  89--92.

\bibitem[\protect\citeauthoryear{Seo, Kembhavi, Farhadi, and Hajishirzi}{Seo
  et~al.}{2016}]{seo2016bidirectional}
Seo, M.~J., A.~Kembhavi, A.~Farhadi, and H.~Hajishirzi 2016.
\newblock Bidirectional attention flow for machine comprehension.
\newblock {\em CoRR\/}~{\em abs/1611.01603}.

\bibitem[\protect\citeauthoryear{Surdeanu}{Surdeanu}{2013}]{surdeanu2013overview}
Surdeanu, M. 2013.
\newblock Overview of the {TAC2013} knowledge base population evaluation:
  English slot filling and temporal slot filling.
\newblock In {\em Proceedings of the Sixth Text Analysis Conference, {TAC} 2013
  \url{https://tac.nist.gov/publications/2013/papers.html}}, Gaithersburg,
  Maryland, USA.

\bibitem[\protect\citeauthoryear{Vaswani, Shazeer, Parmar, Uszkoreit, Jones,
  Gomez, Kaiser, and Polosukhin}{Vaswani et~al.}{2017}]{vaswani2017attention}
Vaswani, A., N.~Shazeer, N.~Parmar, J.~Uszkoreit, L.~Jones, A.~N. Gomez,
  L.~Kaiser, and I.~Polosukhin 2017.
\newblock Attention is all you need.
\newblock {\em CoRR\/}~{\em abs/1706.03762}.

\bibitem[\protect\citeauthoryear{Verga, Belanger, Strubell, Roth, and
  McCallum}{Verga et~al.}{2016}]{verga2015multilingual}
Verga, P., D.~Belanger, E.~Strubell, B.~Roth, and A.~McCallum 2016.
\newblock Multilingual relation extraction using compositional universal
  schema.
\newblock In {\em {NAACL} {HLT} 2016, The 2016 Conference of the North American
  Chapter of the Association for Computational Linguistics: Human Language
  Technologies}, San Diego, California, USA, pp.\  886--96.

\bibitem[\protect\citeauthoryear{Vinyals, Fortunato, and Jaitly}{Vinyals
  et~al.}{2015}]{DBLP:conf/nips/VinyalsFJ15}
Vinyals, O., M.~Fortunato, and N.~Jaitly 2015.
\newblock Pointer networks.
\newblock In C.~Cortes, N.~D. Lawrence, D.~D. Lee, M.~Sugiyama, and R.~Garnett
  (Eds.), {\em Advances in Neural Information Processing Systems 28: Annual
  Conference on Neural Information Processing Systems}, Montreal, Quebec,
  Canada, pp.\  2692--700.

\bibitem[\protect\citeauthoryear{Vu, Adel, Gupta, and Sch{\"{u}}tze}{Vu
  et~al.}{2016}]{vu2016combining}
Vu, N.~T., H.~Adel, P.~Gupta, and H.~Sch{\"{u}}tze 2016.
\newblock Combining recurrent and convolutional neural networks for relation
  classification.
\newblock In {\em {NAACL} {HLT} 2016, The 2016 Conference of the North American
  Chapter of the Association for Computational Linguistics: Human Language
  Technologies}, San Diego, California, USA, pp.\  534--9.

\bibitem[\protect\citeauthoryear{Wang, Yang, Wei, Chang, and Zhou}{Wang
  et~al.}{2017}]{wang2017rnet}
Wang, W., N.~Yang, F.~Wei, B.~Chang, and M.~Zhou 2017.
\newblock Gated self-matching networks for reading comprehension and question
  answering.
\newblock In {\em Proceedings of the 55th Annual Meeting of the Association for
  Computational Linguistics (Volume 1: Long Papers)}, Vancouver, Canada, pp.\
  189--98.

\bibitem[\protect\citeauthoryear{Xu, Jia, Mou, Li, Chen, Lu, and Jin}{Xu
  et~al.}{2016}]{xu2016improved}
Xu, Y., R.~Jia, L.~Mou, G.~Li, Y.~Chen, Y.~Lu, and Z.~Jin 2016.
\newblock Improved relation classification by deep recurrent neural networks
  with data augmentation.
\newblock In {\em Proceedings of COLING 2016, the 26th International Conference
  on Computational Linguistics: Technical Papers}, Osaka, Japan, pp.\
  1461--70.

\bibitem[\protect\citeauthoryear{Zeng, Liu, Lai, Zhou, and Zhao}{Zeng
  et~al.}{2014}]{zeng2014relation}
Zeng, D., K.~Liu, S.~Lai, G.~Zhou, and J.~Zhao 2014.
\newblock Relation classification via convolutional deep neural network.
\newblock In {\em Proceedings of COLING 2014, the 25th International Conference
  on Computational Linguistics: Technical Papers}, Dublin, Ireland, pp.\
  2335--44.

\bibitem[\protect\citeauthoryear{Zhang, Chaganty, Paranjape, Chen, Bolton, Qi,
  and Manning}{Zhang et~al.}{2016}]{zhang2016stanford}
Zhang, Y., A.~Chaganty, A.~Paranjape, D.~Chen, J.~Bolton, P.~Qi, and C.~D.
  Manning 2016.
\newblock Stanford at tac kbp 2016: Sealing pipeline leaks and understanding
  chinese.
\newblock {\em Proceedings of TAC\/}.

\bibitem[\protect\citeauthoryear{Zhang, Zhong, Chen, Angeli, and Manning}{Zhang
  et~al.}{2017}]{zhang2017position}
Zhang, Y., V.~Zhong, D.~Chen, G.~Angeli, and C.~D. Manning 2017.
\newblock Position-aware attention and supervised data improve slot filling.
\newblock In {\em Proceedings of the 2017 Conference on Empirical Methods in
  Natural Language Processing}, pp.\  35--45.

\bibitem[\protect\citeauthoryear{Zheng, Wang, Bao, Hao, Zhou, and Xu}{Zheng
  et~al.}{2017}]{zheng2017joint}
Zheng, S., F.~Wang, H.~Bao, Y.~Hao, P.~Zhou, and B.~Xu 2017.
\newblock Joint extraction of entities and relations based on a novel tagging
  scheme.
\newblock In {\em Proceedings of the 55th Annual Meeting of the Association for
  Computational Linguistics (Volume 1: Long Papers)}, Vancouver, Canada, pp.\
  1227--36.

\end{thebibliography}


\begin{thebibliography}{}

\bibitem [\protect \citeauthoryear {%
Adel%
, Roth%
\BCBL {}\ \BBA {} Sch{\"{u}}tze%
}{%
Adel%
\ \protect \BOthers {.}}{%
{\protect \APACyear {2016}}%
}]{%
adel2016comparing}
\APACinsertmetastar {%
adel2016comparing}%
\begin{APACrefauthors}%
Adel, H.%
, Roth, B.%
\BCBL {}\ \BBA {} Sch{\"{u}}tze, H.%
\end{APACrefauthors}%
\unskip\
\newblock
\APACrefYearMonthDay{2016}{June}{}.
\newblock
{\BBOQ}\APACrefatitle {Comparing Convolutional Neural Networks to Traditional
  Models for Slot Filling} {Comparing convolutional neural networks to
  traditional models for slot filling}.{\BBCQ}
\newblock
\BIn{} \APACrefbtitle {{NAACL} {HLT} 2016, The 2016 Conference of the North
  American Chapter of the Association for Computational Linguistics: Human
  Language Technologies} {{NAACL} {HLT} 2016, the 2016 conference of the north
  american chapter of the association for computational linguistics: Human
  language technologies}\ (\BPGS\ 828--838).
\newblock
\APACaddressPublisher{San Diego, California, USA}{}.
\PrintBackRefs{\CurrentBib}

\bibitem [\protect \citeauthoryear {%
Ba%
, Kiros%
\BCBL {}\ \BBA {} Hinton%
}{%
Ba%
\ \protect \BOthers {.}}{%
{\protect \APACyear {2016}}%
}]{%
ba2016layer}
\APACinsertmetastar {%
ba2016layer}%
\begin{APACrefauthors}%
Ba, J.%
, Kiros, R.%
\BCBL {}\ \BBA {} Hinton, G\BPBI E.%
\end{APACrefauthors}%
\unskip\
\newblock
\APACrefYearMonthDay{2016}{}{}.
\newblock
{\BBOQ}\APACrefatitle {Layer Normalization} {Layer normalization}.{\BBCQ}
\newblock
\APACjournalVolNumPages{CoRR}{abs/1607.06450}{}{}.
\PrintBackRefs{\CurrentBib}

\bibitem [\protect \citeauthoryear {%
Bahdanau%
, Cho%
\BCBL {}\ \BBA {} Bengio%
}{%
Bahdanau%
\ \protect \BOthers {.}}{%
{\protect \APACyear {2014}}%
}]{%
DBLP:journals/corr/BahdanauCB14}
\APACinsertmetastar {%
DBLP:journals/corr/BahdanauCB14}%
\begin{APACrefauthors}%
Bahdanau, D.%
, Cho, K.%
\BCBL {}\ \BBA {} Bengio, Y.%
\end{APACrefauthors}%
\unskip\
\newblock
\APACrefYearMonthDay{2014}{}{}.
\newblock
{\BBOQ}\APACrefatitle {Neural Machine Translation by Jointly Learning to Align
  and Translate} {Neural machine translation by jointly learning to align and
  translate}.{\BBCQ}
\newblock
\APACjournalVolNumPages{CoRR}{abs/1409.0473}{}{}.
\newblock
\begin{APACrefURL} \url{http://arxiv.org/abs/1409.0473} \end{APACrefURL}
\PrintBackRefs{\CurrentBib}

\bibitem [\protect \citeauthoryear {%
Chen%
, Fisch%
, Weston%
\BCBL {}\ \BBA {} Bordes%
}{%
Chen%
\ \protect \BOthers {.}}{%
{\protect \APACyear {2017}}%
}]{%
chen2017reading}
\APACinsertmetastar {%
chen2017reading}%
\begin{APACrefauthors}%
Chen, D.%
, Fisch, A.%
, Weston, J.%
\BCBL {}\ \BBA {} Bordes, A.%
\end{APACrefauthors}%
\unskip\
\newblock
\APACrefYearMonthDay{2017}{July}{}.
\newblock
{\BBOQ}\APACrefatitle {Reading Wikipedia to Answer Open-Domain Questions}
  {Reading wikipedia to answer open-domain questions}.{\BBCQ}
\newblock
\BIn{} \APACrefbtitle {Proceedings of the 55th Annual Meeting of the
  Association for Computational Linguistics (Volume 1: Long Papers)}
  {Proceedings of the 55th annual meeting of the association for computational
  linguistics (volume 1: Long papers)}\ (\BPGS\ 1870--1879).
\newblock
\APACaddressPublisher{Vancouver, Canada}{Association for Computational
  Linguistics}.
\PrintBackRefs{\CurrentBib}

\bibitem [\protect \citeauthoryear {%
Chinchor%
\ \BBA {} Robinson%
}{%
Chinchor%
\ \BBA {} Robinson%
}{%
{\protect \APACyear {1997}}%
}]{%
chinchor1997muc}
\APACinsertmetastar {%
chinchor1997muc}%
\begin{APACrefauthors}%
Chinchor, N.%
\BCBT {}\ \BBA {} Robinson, P.%
\end{APACrefauthors}%
\unskip\
\newblock
\APACrefYearMonthDay{1997}{}{}.
\newblock
{\BBOQ}\APACrefatitle {MUC-7 named entity task definition} {Muc-7 named entity
  task definition}.{\BBCQ}
\newblock
\BIn{} \APACrefbtitle {Proceedings of the 7th Conference on Message
  Understanding} {Proceedings of the 7th conference on message understanding}\
  (\BVOL~29).
\PrintBackRefs{\CurrentBib}

\bibitem [\protect \citeauthoryear {%
Choi%
, Breck%
\BCBL {}\ \BBA {} Cardie%
}{%
Choi%
\ \protect \BOthers {.}}{%
{\protect \APACyear {2006}}%
}]{%
choi2006joint}
\APACinsertmetastar {%
choi2006joint}%
\begin{APACrefauthors}%
Choi, Y.%
, Breck, E.%
\BCBL {}\ \BBA {} Cardie, C.%
\end{APACrefauthors}%
\unskip\
\newblock
\APACrefYearMonthDay{2006}{July}{}.
\newblock
{\BBOQ}\APACrefatitle {Joint extraction of entities and relations for opinion
  recognition} {Joint extraction of entities and relations for opinion
  recognition}.{\BBCQ}
\newblock
\BIn{} \APACrefbtitle {Proceedings of the 2006 Conference on Empirical Methods
  in Natural Language Processing (EMNLP)} {Proceedings of the 2006 conference
  on empirical methods in natural language processing (emnlp)}\ (\BPGS\
  431--439).
\newblock
\APACaddressPublisher{Sydney, Australia}{}.
\PrintBackRefs{\CurrentBib}

\bibitem [\protect \citeauthoryear {%
Chung%
, Çaglar G\"{u}lçehre%
, Cho%
\BCBL {}\ \BBA {} Bengio%
}{%
Chung%
\ \protect \BOthers {.}}{%
{\protect \APACyear {2014}}%
}]{%
chung2014empirical}
\APACinsertmetastar {%
chung2014empirical}%
\begin{APACrefauthors}%
Chung, J.%
, Çaglar G\"{u}lçehre%
, Cho, K.%
\BCBL {}\ \BBA {} Bengio, Y.%
\end{APACrefauthors}%
\unskip\
\newblock
\APACrefYearMonthDay{2014}{}{}.
\newblock
{\BBOQ}\APACrefatitle {Empirical Evaluation of Gated Recurrent Neural Networks
  on Sequence Modeling} {Empirical evaluation of gated recurrent neural
  networks on sequence modeling}.{\BBCQ}
\newblock
\APACjournalVolNumPages{CoRR}{abs/1412.3555}{}{}.
\PrintBackRefs{\CurrentBib}

\bibitem [\protect \citeauthoryear {%
Collobert%
\ \protect \BOthers {.}}{%
Collobert%
\ \protect \BOthers {.}}{%
{\protect \APACyear {2011}}%
}]{%
collobert2011natural}
\APACinsertmetastar {%
collobert2011natural}%
\begin{APACrefauthors}%
Collobert, R.%
, Weston, J.%
, Bottou, L.%
, Karlen, M.%
, Kavukcuoglu, K.%
\BCBL {}\ \BBA {} Kuksa, P\BPBI P.%
\end{APACrefauthors}%
\unskip\
\newblock
\APACrefYearMonthDay{2011}{}{}.
\newblock
{\BBOQ}\APACrefatitle {Natural Language Processing (Almost) from Scratch}
  {Natural language processing (almost) from scratch}.{\BBCQ}
\newblock
\APACjournalVolNumPages{Journal of Machine Learning
  Research}{12}{}{2493--2537}.
\PrintBackRefs{\CurrentBib}

\bibitem [\protect \citeauthoryear {%
Culotta%
, McCallum%
\BCBL {}\ \BBA {} Betz%
}{%
Culotta%
\ \protect \BOthers {.}}{%
{\protect \APACyear {2006}}%
}]{%
culotta2006integrating}
\APACinsertmetastar {%
culotta2006integrating}%
\begin{APACrefauthors}%
Culotta, A.%
, McCallum, A.%
\BCBL {}\ \BBA {} Betz, J.%
\end{APACrefauthors}%
\unskip\
\newblock
\APACrefYearMonthDay{2006}{June}{}.
\newblock
{\BBOQ}\APACrefatitle {Integrating probabilistic extraction models and data
  mining to discover relations and patterns in text} {Integrating probabilistic
  extraction models and data mining to discover relations and patterns in
  text}.{\BBCQ}
\newblock
\BIn{} \APACrefbtitle {{NAACL} {HLT} 2006, Proceedings of the Human Language
  Technology Conference of the North American Chapter of the Association of
  Computational Linguistics} {{NAACL} {HLT} 2006, proceedings of the human
  language technology conference of the north american chapter of the
  association of computational linguistics}\ (\BPGS\ 296--303).
\newblock
\APACaddressPublisher{New York, USA}{}.
\PrintBackRefs{\CurrentBib}

\bibitem [\protect \citeauthoryear {%
dos Santos%
, Xiang%
\BCBL {}\ \BBA {} Zhou%
}{%
dos Santos%
\ \protect \BOthers {.}}{%
{\protect \APACyear {2015}}%
}]{%
dossantos2015classifying}
\APACinsertmetastar {%
dossantos2015classifying}%
\begin{APACrefauthors}%
dos Santos, C\BPBI N.%
, Xiang, B.%
\BCBL {}\ \BBA {} Zhou, B.%
\end{APACrefauthors}%
\unskip\
\newblock
\APACrefYearMonthDay{2015}{July}{}.
\newblock
{\BBOQ}\APACrefatitle {Classifying Relations by Ranking with Convolutional
  Neural Networks} {Classifying relations by ranking with convolutional neural
  networks}.{\BBCQ}
\newblock
\BIn{} \APACrefbtitle {Proceedings of the 53rd Annual Meeting of the
  Association for Computational Linguistics and the 7th International Joint
  Conference on Natural Language Processing of the Asian Federation of Natural
  Language Processing} {Proceedings of the 53rd annual meeting of the
  association for computational linguistics and the 7th international joint
  conference on natural language processing of the asian federation of natural
  language processing}\ (\BVOL~1, \BPGS\ 626--634).
\newblock
\APACaddressPublisher{Beijing, China}{}.
\PrintBackRefs{\CurrentBib}

\bibitem [\protect \citeauthoryear {%
G{\"u}l{\c c}ehre%
, Ahn%
, Nallapati%
, Zhou%
\BCBL {}\ \BBA {} Bengio%
}{%
G{\"u}l{\c c}ehre%
\ \protect \BOthers {.}}{%
{\protect \APACyear {2016}}%
}]{%
gulcehre-EtAl:2016:P16-1}
\APACinsertmetastar {%
gulcehre-EtAl:2016:P16-1}%
\begin{APACrefauthors}%
G{\"u}l{\c c}ehre, {\c C}.%
, Ahn, S.%
, Nallapati, R.%
, Zhou, B.%
\BCBL {}\ \BBA {} Bengio, Y.%
\end{APACrefauthors}%
\unskip\
\newblock
\APACrefYearMonthDay{2016}{August}{}.
\newblock
{\BBOQ}\APACrefatitle {Pointing the Unknown Words} {Pointing the unknown
  words}.{\BBCQ}
\newblock
\BIn{} \APACrefbtitle {Proceedings of the 54th Annual Meeting of the
  Association for Computational Linguistics} {Proceedings of the 54th annual
  meeting of the association for computational linguistics}\ (\BVOL~1, \BPGS\
  140--149).
\newblock
\APACaddressPublisher{Berlin, Germany}{}.
\newblock
\begin{APACrefURL} \url{http://www.aclweb.org/anthology/P16-1014}
  \end{APACrefURL}
\PrintBackRefs{\CurrentBib}

\bibitem [\protect \citeauthoryear {%
Gupta%
, Sch{\"u}tze%
\BCBL {}\ \BBA {} Andrassy%
}{%
Gupta%
\ \protect \BOthers {.}}{%
{\protect \APACyear {2016}}%
}]{%
gupta2016table}
\APACinsertmetastar {%
gupta2016table}%
\begin{APACrefauthors}%
Gupta, P.%
, Sch{\"u}tze, H.%
\BCBL {}\ \BBA {} Andrassy, B.%
\end{APACrefauthors}%
\unskip\
\newblock
\APACrefYearMonthDay{2016}{December}{}.
\newblock
{\BBOQ}\APACrefatitle {Table Filling Multi-Task Recurrent Neural Network for
  Joint Entity and Relation Extraction} {Table filling multi-task recurrent
  neural network for joint entity and relation extraction}.{\BBCQ}
\newblock
\BIn{} \APACrefbtitle {Proceedings of COLING 2016, the 26th International
  Conference on Computational Linguistics: Technical Papers} {Proceedings of
  coling 2016, the 26th international conference on computational linguistics:
  Technical papers}\ (\BPGS\ 2537--2547).
\newblock
\APACaddressPublisher{Osaka, Japan}{}.
\PrintBackRefs{\CurrentBib}

\bibitem [\protect \citeauthoryear {%
Hewlett%
\ \protect \BOthers {.}}{%
Hewlett%
\ \protect \BOthers {.}}{%
{\protect \APACyear {2016}}%
}]{%
hewlett2016wikireading}
\APACinsertmetastar {%
hewlett2016wikireading}%
\begin{APACrefauthors}%
Hewlett, D.%
, Lacoste, A.%
, Jones, L.%
, Polosukhin, I.%
, Fandrianto, A.%
, Han, J.%
\BDBL {}Berthelot, D.%
\end{APACrefauthors}%
\unskip\
\newblock
\APACrefYearMonthDay{2016}{August}{}.
\newblock
{\BBOQ}\APACrefatitle {Wikireading: A novel large-scale language understanding
  task over wikipedia} {Wikireading: A novel large-scale language understanding
  task over wikipedia}.{\BBCQ}
\newblock
\BIn{} \APACrefbtitle {Proceedings of the 54th Annual Meeting of the
  Association for Computational Linguistics} {Proceedings of the 54th annual
  meeting of the association for computational linguistics}\ (\BPGS\
  1535--1545).
\newblock
\APACaddressPublisher{Berlin, Germany}{}.
\PrintBackRefs{\CurrentBib}

\bibitem [\protect \citeauthoryear {%
Hochreiter%
\ \BBA {} Schmidhuber%
}{%
Hochreiter%
\ \BBA {} Schmidhuber%
}{%
{\protect \APACyear {1997}}%
}]{%
hochreiter1997long}
\APACinsertmetastar {%
hochreiter1997long}%
\begin{APACrefauthors}%
Hochreiter, S.%
\BCBT {}\ \BBA {} Schmidhuber, J.%
\end{APACrefauthors}%
\unskip\
\newblock
\APACrefYearMonthDay{1997}{}{}.
\newblock
{\BBOQ}\APACrefatitle {Long short-term memory} {Long short-term memory}.{\BBCQ}
\newblock
\APACjournalVolNumPages{Neural computation}{9}{8}{1735--1780}.
\PrintBackRefs{\CurrentBib}

\bibitem [\protect \citeauthoryear {%
Hoffmann%
, Zhang%
\BCBL {}\ \BBA {} Weld%
}{%
Hoffmann%
\ \protect \BOthers {.}}{%
{\protect \APACyear {2010}}%
}]{%
hoffmann2010learning}
\APACinsertmetastar {%
hoffmann2010learning}%
\begin{APACrefauthors}%
Hoffmann, R.%
, Zhang, C.%
\BCBL {}\ \BBA {} Weld, D\BPBI S.%
\end{APACrefauthors}%
\unskip\
\newblock
\APACrefYearMonthDay{2010}{July}{}.
\newblock
{\BBOQ}\APACrefatitle {Learning 5000 relational extractors} {Learning 5000
  relational extractors}.{\BBCQ}
\newblock
\BIn{} \APACrefbtitle {Proceedings of the 48th Annual Meeting of the
  Association for Computational Linguistics} {Proceedings of the 48th annual
  meeting of the association for computational linguistics}\ (\BPGS\ 286--295).
\newblock
\APACaddressPublisher{Uppsala, Sweden}{}.
\PrintBackRefs{\CurrentBib}

\bibitem [\protect \citeauthoryear {%
Ioffe%
\ \BBA {} Szegedy%
}{%
Ioffe%
\ \BBA {} Szegedy%
}{%
{\protect \APACyear {2015}}%
}]{%
ioffe2015batch}
\APACinsertmetastar {%
ioffe2015batch}%
\begin{APACrefauthors}%
Ioffe, S.%
\BCBT {}\ \BBA {} Szegedy, C.%
\end{APACrefauthors}%
\unskip\
\newblock
\APACrefYearMonthDay{2015}{}{}.
\newblock
{\BBOQ}\APACrefatitle {Batch normalization: Accelerating deep network training
  by reducing internal covariate shift} {Batch normalization: Accelerating deep
  network training by reducing internal covariate shift}.{\BBCQ}
\newblock
\BIn{} \APACrefbtitle {Proceedings of the 32nd International Conference on
  Machine Learning} {Proceedings of the 32nd international conference on
  machine learning}\ (\BVOL~37, \BPGS\ 448--456).
\newblock
\APACaddressPublisher{Lille, France}{}.
\PrintBackRefs{\CurrentBib}

\bibitem [\protect \citeauthoryear {%
Katiyar%
\ \BBA {} Cardie%
}{%
Katiyar%
\ \BBA {} Cardie%
}{%
{\protect \APACyear {2016}}%
}]{%
katiyar2016investigating}
\APACinsertmetastar {%
katiyar2016investigating}%
\begin{APACrefauthors}%
Katiyar, A.%
\BCBT {}\ \BBA {} Cardie, C.%
\end{APACrefauthors}%
\unskip\
\newblock
\APACrefYearMonthDay{2016}{August}{}.
\newblock
{\BBOQ}\APACrefatitle {Investigating {LSTMs} for Joint Extraction of Opinion
  Entities and Relations.} {Investigating {LSTMs} for joint extraction of
  opinion entities and relations.}{\BBCQ}
\newblock
\BIn{} \APACrefbtitle {Proceedings of the 54th Annual Meeting of the
  Association for Computational Linguistics} {Proceedings of the 54th annual
  meeting of the association for computational linguistics}\ (\BPGS\ 919--929).
\newblock
\APACaddressPublisher{Berlin, Germany}{}.
\PrintBackRefs{\CurrentBib}

\bibitem [\protect \citeauthoryear {%
Lample%
, Ballesteros%
, Subramanian%
, Kawakami%
\BCBL {}\ \BBA {} Dyer%
}{%
Lample%
\ \protect \BOthers {.}}{%
{\protect \APACyear {2016}}%
}]{%
lample2016neural}
\APACinsertmetastar {%
lample2016neural}%
\begin{APACrefauthors}%
Lample, G.%
, Ballesteros, M.%
, Subramanian, S.%
, Kawakami, K.%
\BCBL {}\ \BBA {} Dyer, C.%
\end{APACrefauthors}%
\unskip\
\newblock
\APACrefYearMonthDay{2016}{June}{}.
\newblock
{\BBOQ}\APACrefatitle {Neural architectures for named entity recognition}
  {Neural architectures for named entity recognition}.{\BBCQ}
\newblock
\BIn{} \APACrefbtitle {{NAACL} {HLT} 2016, The 2016 Conference of the North
  American Chapter of the Association for Computational Linguistics: Human
  Language Technologies} {{NAACL} {HLT} 2016, the 2016 conference of the north
  american chapter of the association for computational linguistics: Human
  language technologies}\ (\BPGS\ 260--270).
\newblock
\APACaddressPublisher{San Diego, California, USA}{}.
\PrintBackRefs{\CurrentBib}

\bibitem [\protect \citeauthoryear {%
Levy%
, Seo%
, Choi%
\BCBL {}\ \BBA {} Zettlemoyer%
}{%
Levy%
\ \protect \BOthers {.}}{%
{\protect \APACyear {2017}}%
}]{%
levy2017zero}
\APACinsertmetastar {%
levy2017zero}%
\begin{APACrefauthors}%
Levy, O.%
, Seo, M.%
, Choi, E.%
\BCBL {}\ \BBA {} Zettlemoyer, L.%
\end{APACrefauthors}%
\unskip\
\newblock
\APACrefYearMonthDay{2017}{August}{}.
\newblock
{\BBOQ}\APACrefatitle {Zero-Shot Relation Extraction via Reading Comprehension}
  {Zero-shot relation extraction via reading comprehension}.{\BBCQ}
\newblock
\BIn{} \APACrefbtitle {Proceedings of the 21st Conference on Computational
  Natural Language Learning (CoNLL 2017)} {Proceedings of the 21st conference
  on computational natural language learning (conll 2017)}\ (\BPGS\ 333--342).
\newblock
\APACaddressPublisher{Vancouver, Canada}{Association for Computational
  Linguistics}.
\PrintBackRefs{\CurrentBib}

\bibitem [\protect \citeauthoryear {%
Mintz%
, Bills%
, Snow%
\BCBL {}\ \BBA {} Jurafsky%
}{%
Mintz%
\ \protect \BOthers {.}}{%
{\protect \APACyear {2009}}%
}]{%
mintz2009distant}
\APACinsertmetastar {%
mintz2009distant}%
\begin{APACrefauthors}%
Mintz, M.%
, Bills, S.%
, Snow, R.%
\BCBL {}\ \BBA {} Jurafsky, D.%
\end{APACrefauthors}%
\unskip\
\newblock
\APACrefYearMonthDay{2009}{August}{}.
\newblock
{\BBOQ}\APACrefatitle {Distant supervision for relation extraction without
  labeled data} {Distant supervision for relation extraction without labeled
  data}.{\BBCQ}
\newblock
\BIn{} \APACrefbtitle {Proceedings of the Joint Conference of the 47th Annual
  Meeting of the ACL and the 4th International Joint Conference on Natural
  Language Processing of the AFNLP} {Proceedings of the joint conference of the
  47th annual meeting of the acl and the 4th international joint conference on
  natural language processing of the afnlp}\ (\BVOL~2, \BPGS\ 1003--1011).
\newblock
\APACaddressPublisher{Suntec, Singapore}{}.
\PrintBackRefs{\CurrentBib}

\bibitem [\protect \citeauthoryear {%
Miwa%
\ \BBA {} Bansal%
}{%
Miwa%
\ \BBA {} Bansal%
}{%
{\protect \APACyear {2016}}%
}]{%
miwa2016end}
\APACinsertmetastar {%
miwa2016end}%
\begin{APACrefauthors}%
Miwa, M.%
\BCBT {}\ \BBA {} Bansal, M.%
\end{APACrefauthors}%
\unskip\
\newblock
\APACrefYearMonthDay{2016}{August}{}.
\newblock
{\BBOQ}\APACrefatitle {End-to-end relation extraction using {LSTMs} on
  sequences and tree structures} {End-to-end relation extraction using {LSTMs}
  on sequences and tree structures}.{\BBCQ}
\newblock
\BIn{} \APACrefbtitle {Proceedings of the 54th Annual Meeting of the
  Association for Computational Linguistics} {Proceedings of the 54th annual
  meeting of the association for computational linguistics}\ (\BPGS\
  1105--1116).
\newblock
\APACaddressPublisher{Berlin, Germany}{}.
\PrintBackRefs{\CurrentBib}

\bibitem [\protect \citeauthoryear {%
Miwa%
\ \BBA {} Sasaki%
}{%
Miwa%
\ \BBA {} Sasaki%
}{%
{\protect \APACyear {2014}}%
}]{%
miwa2014modeling}
\APACinsertmetastar {%
miwa2014modeling}%
\begin{APACrefauthors}%
Miwa, M.%
\BCBT {}\ \BBA {} Sasaki, Y.%
\end{APACrefauthors}%
\unskip\
\newblock
\APACrefYearMonthDay{2014}{October}{}.
\newblock
{\BBOQ}\APACrefatitle {Modeling Joint Entity and Relation Extraction with Table
  Representation.} {Modeling joint entity and relation extraction with table
  representation.}{\BBCQ}
\newblock
\BIn{} \APACrefbtitle {Proceedings of the 2014 Conference on Empirical Methods
  in Natural Language Processing (EMNLP)} {Proceedings of the 2014 conference
  on empirical methods in natural language processing (emnlp)}\ (\BPGS\
  1858--1869).
\newblock
\APACaddressPublisher{Doha, Qatar}{}.
\PrintBackRefs{\CurrentBib}

\bibitem [\protect \citeauthoryear {%
Pink%
, Nothman%
\BCBL {}\ \BBA {} Curran%
}{%
Pink%
\ \protect \BOthers {.}}{%
{\protect \APACyear {2014}}%
}]{%
pink2014analysing}
\APACinsertmetastar {%
pink2014analysing}%
\begin{APACrefauthors}%
Pink, G.%
, Nothman, J.%
\BCBL {}\ \BBA {} Curran, J\BPBI R.%
\end{APACrefauthors}%
\unskip\
\newblock
\APACrefYearMonthDay{2014}{October}{}.
\newblock
{\BBOQ}\APACrefatitle {Analysing recall loss in named entity slot filling}
  {Analysing recall loss in named entity slot filling}.{\BBCQ}
\newblock
\BIn{} \APACrefbtitle {Proceedings of the 2014 Conference on Empirical Methods
  in Natural Language Processing, {EMNLP} 2014} {Proceedings of the 2014
  conference on empirical methods in natural language processing, {EMNLP}
  2014}\ (\BPGS\ 820--830).
\newblock
\APACaddressPublisher{Doha, Qatar}{}.
\PrintBackRefs{\CurrentBib}

\bibitem [\protect \citeauthoryear {%
Rajpurkar%
, Zhang%
, Lopyrev%
\BCBL {}\ \BBA {} Liang%
}{%
Rajpurkar%
\ \protect \BOthers {.}}{%
{\protect \APACyear {2016}}%
}]{%
rajpurkar-EtAl:2016:EMNLP2016}
\APACinsertmetastar {%
rajpurkar-EtAl:2016:EMNLP2016}%
\begin{APACrefauthors}%
Rajpurkar, P.%
, Zhang, J.%
, Lopyrev, K.%
\BCBL {}\ \BBA {} Liang, P.%
\end{APACrefauthors}%
\unskip\
\newblock
\APACrefYearMonthDay{2016}{November}{}.
\newblock
{\BBOQ}\APACrefatitle {SQuAD: 100,000+ Questions for Machine Comprehension of
  Text} {Squad: 100,000+ questions for machine comprehension of text}.{\BBCQ}
\newblock
\BIn{} \APACrefbtitle {Proceedings of the 2016 Conference on Empirical Methods
  in Natural Language Processing} {Proceedings of the 2016 conference on
  empirical methods in natural language processing}\ (\BPGS\ 2383--2392).
\newblock
\APACaddressPublisher{Austin, Texas}{Association for Computational
  Linguistics}.
\newblock
\begin{APACrefURL} \url{https://aclweb.org/anthology/D16-1264} \end{APACrefURL}
\PrintBackRefs{\CurrentBib}

\bibitem [\protect \citeauthoryear {%
Ren%
\ \protect \BOthers {.}}{%
Ren%
\ \protect \BOthers {.}}{%
{\protect \APACyear {2017}}%
}]{%
ren2017cotype}
\APACinsertmetastar {%
ren2017cotype}%
\begin{APACrefauthors}%
Ren, X.%
, Wu, Z.%
, He, W.%
, Qu, M.%
, Voss, C\BPBI R.%
, Ji, H.%
\BDBL {}Han, J.%
\end{APACrefauthors}%
\unskip\
\newblock
\APACrefYearMonthDay{2017}{}{}.
\newblock
{\BBOQ}\APACrefatitle {CoType: Joint extraction of typed entities and relations
  with knowledge bases} {Cotype: Joint extraction of typed entities and
  relations with knowledge bases}.{\BBCQ}
\newblock
\BIn{} \APACrefbtitle {Proceedings of the 26th International Conference on
  World Wide Web} {Proceedings of the 26th international conference on world
  wide web}\ (\BPGS\ 1015--1024).
\newblock
\APACaddressPublisher{Perth, Australia}{}.
\PrintBackRefs{\CurrentBib}

\bibitem [\protect \citeauthoryear {%
Roth%
, Barth%
, Chrupala%
, Gropp%
\BCBL {}\ \BBA {} Klakow%
}{%
Roth%
\ \protect \BOthers {.}}{%
{\protect \APACyear {2014}}%
}]{%
roth2014relationfactory}
\APACinsertmetastar {%
roth2014relationfactory}%
\begin{APACrefauthors}%
Roth, B.%
, Barth, T.%
, Chrupala, G.%
, Gropp, M.%
\BCBL {}\ \BBA {} Klakow, D.%
\end{APACrefauthors}%
\unskip\
\newblock
\APACrefYearMonthDay{2014}{April}{}.
\newblock
{\BBOQ}\APACrefatitle {{RelationFactory}: A Fast, Modular and Effective System
  for Knowledge Base Population.} {{RelationFactory}: A fast, modular and
  effective system for knowledge base population.}{\BBCQ}
\newblock
\BIn{} \APACrefbtitle {Proceedings of the Demonstrations at the 14th Conference
  of the European Chapter of the Association for Computational Linguistics}
  {Proceedings of the demonstrations at the 14th conference of the european
  chapter of the association for computational linguistics}\ (\BPGS\ 89--92).
\newblock
\APACaddressPublisher{Gothenburg, Sweden}{}.
\PrintBackRefs{\CurrentBib}

\bibitem [\protect \citeauthoryear {%
Seo%
, Kembhavi%
, Farhadi%
\BCBL {}\ \BBA {} Hajishirzi%
}{%
Seo%
\ \protect \BOthers {.}}{%
{\protect \APACyear {2016}}%
}]{%
seo2016bidirectional}
\APACinsertmetastar {%
seo2016bidirectional}%
\begin{APACrefauthors}%
Seo, M.%
, Kembhavi, A.%
, Farhadi, A.%
\BCBL {}\ \BBA {} Hajishirzi, H.%
\end{APACrefauthors}%
\unskip\
\newblock
\APACrefYearMonthDay{2016}{}{}.
\newblock
{\BBOQ}\APACrefatitle {Bidirectional attention flow for machine comprehension}
  {Bidirectional attention flow for machine comprehension}.{\BBCQ}
\newblock
\APACjournalVolNumPages{arXiv preprint arXiv:1611.01603}{}{}{}.
\PrintBackRefs{\CurrentBib}

\bibitem [\protect \citeauthoryear {%
Surdeanu%
}{%
Surdeanu%
}{%
{\protect \APACyear {2013}}%
}]{%
surdeanu2013overview}
\APACinsertmetastar {%
surdeanu2013overview}%
\begin{APACrefauthors}%
Surdeanu, M.%
\end{APACrefauthors}%
\unskip\
\newblock
\APACrefYearMonthDay{2013}{November}{}.
\newblock
{\BBOQ}\APACrefatitle {Overview of the {TAC2013} Knowledge Base Population
  Evaluation: English Slot Filling and Temporal Slot Filling} {Overview of the
  {TAC2013} knowledge base population evaluation: English slot filling and
  temporal slot filling}.{\BBCQ}
\newblock
\BIn{} \APACrefbtitle {Proceedings of the Sixth Text Analysis Conference, {TAC}
  2013.} {Proceedings of the sixth text analysis conference, {TAC} 2013.}
\newblock
\APACaddressPublisher{Gaithersburg, Maryland, USA}{}.
\PrintBackRefs{\CurrentBib}

\bibitem [\protect \citeauthoryear {%
Vaswani%
\ \protect \BOthers {.}}{%
Vaswani%
\ \protect \BOthers {.}}{%
{\protect \APACyear {2017}}%
}]{%
vaswani2017attention}
\APACinsertmetastar {%
vaswani2017attention}%
\begin{APACrefauthors}%
Vaswani, A.%
, Shazeer, N.%
, Parmar, N.%
, Uszkoreit, J.%
, Jones, L.%
, Gomez, A\BPBI N.%
\BDBL {}Polosukhin, I.%
\end{APACrefauthors}%
\unskip\
\newblock
\APACrefYearMonthDay{2017}{}{}.
\newblock
{\BBOQ}\APACrefatitle {Attention Is All You Need} {Attention is all you
  need}.{\BBCQ}
\newblock
\APACjournalVolNumPages{arXiv preprint arXiv:1706.03762}{}{}{}.
\PrintBackRefs{\CurrentBib}

\bibitem [\protect \citeauthoryear {%
Verga%
, Belanger%
, Strubell%
, Roth%
\BCBL {}\ \BBA {} McCallum%
}{%
Verga%
\ \protect \BOthers {.}}{%
{\protect \APACyear {2016}}%
}]{%
verga2015multilingual}
\APACinsertmetastar {%
verga2015multilingual}%
\begin{APACrefauthors}%
Verga, P.%
, Belanger, D.%
, Strubell, E.%
, Roth, B.%
\BCBL {}\ \BBA {} McCallum, A.%
\end{APACrefauthors}%
\unskip\
\newblock
\APACrefYearMonthDay{2016}{June}{}.
\newblock
{\BBOQ}\APACrefatitle {Multilingual relation extraction using compositional
  universal schema} {Multilingual relation extraction using compositional
  universal schema}.{\BBCQ}
\newblock
\BIn{} \APACrefbtitle {{NAACL} {HLT} 2016, The 2016 Conference of the North
  American Chapter of the Association for Computational Linguistics: Human
  Language Technologies} {{NAACL} {HLT} 2016, the 2016 conference of the north
  american chapter of the association for computational linguistics: Human
  language technologies}\ (\BPGS\ 886--896).
\newblock
\APACaddressPublisher{San Diego, California, USA}{}.
\PrintBackRefs{\CurrentBib}

\bibitem [\protect \citeauthoryear {%
Vinyals%
, Fortunato%
\BCBL {}\ \BBA {} Jaitly%
}{%
Vinyals%
\ \protect \BOthers {.}}{%
{\protect \APACyear {2015}}%
}]{%
DBLP:conf/nips/VinyalsFJ15}
\APACinsertmetastar {%
DBLP:conf/nips/VinyalsFJ15}%
\begin{APACrefauthors}%
Vinyals, O.%
, Fortunato, M.%
\BCBL {}\ \BBA {} Jaitly, N.%
\end{APACrefauthors}%
\unskip\
\newblock
\APACrefYearMonthDay{2015}{December}{}.
\newblock
{\BBOQ}\APACrefatitle {Pointer Networks} {Pointer networks}.{\BBCQ}
\newblock
\BIn{} C.~Cortes, N\BPBI D.~Lawrence, D\BPBI D.~Lee, M.~Sugiyama\BCBL {}\ \BBA
  {} R.~Garnett\ (\BEDS), \APACrefbtitle {Advances in Neural Information
  Processing Systems 28: Annual Conference on Neural Information Processing
  Systems} {Advances in neural information processing systems 28: Annual
  conference on neural information processing systems}\ (\BPGS\ 2692--2700).
\newblock
\APACaddressPublisher{Montreal, Quebec, Canada}{}.
\newblock
\begin{APACrefURL} \url{http://papers.nips.cc/paper/5866-pointer-networks}
  \end{APACrefURL}
\PrintBackRefs{\CurrentBib}

\bibitem [\protect \citeauthoryear {%
Vu%
, Adel%
, Gupta%
\BCBL {}\ \BBA {} Sch{\"{u}}tze%
}{%
Vu%
\ \protect \BOthers {.}}{%
{\protect \APACyear {2016}}%
}]{%
vu2016combining}
\APACinsertmetastar {%
vu2016combining}%
\begin{APACrefauthors}%
Vu, N\BPBI T.%
, Adel, H.%
, Gupta, P.%
\BCBL {}\ \BBA {} Sch{\"{u}}tze, H.%
\end{APACrefauthors}%
\unskip\
\newblock
\APACrefYearMonthDay{2016}{June}{}.
\newblock
{\BBOQ}\APACrefatitle {Combining Recurrent and Convolutional Neural Networks
  for Relation Classification} {Combining recurrent and convolutional neural
  networks for relation classification}.{\BBCQ}
\newblock
\BIn{} \APACrefbtitle {{NAACL} {HLT} 2016, The 2016 Conference of the North
  American Chapter of the Association for Computational Linguistics: Human
  Language Technologies} {{NAACL} {HLT} 2016, the 2016 conference of the north
  american chapter of the association for computational linguistics: Human
  language technologies}\ (\BPGS\ 534--539).
\newblock
\APACaddressPublisher{San Diego, California, USA}{}.
\PrintBackRefs{\CurrentBib}

\bibitem [\protect \citeauthoryear {%
Wang%
, Yang%
, Wei%
, Chang%
\BCBL {}\ \BBA {} Zhou%
}{%
Wang%
\ \protect \BOthers {.}}{%
{\protect \APACyear {2017}}%
}]{%
wang2017rnet}
\APACinsertmetastar {%
wang2017rnet}%
\begin{APACrefauthors}%
Wang, W.%
, Yang, N.%
, Wei, F.%
, Chang, B.%
\BCBL {}\ \BBA {} Zhou, M.%
\end{APACrefauthors}%
\unskip\
\newblock
\APACrefYearMonthDay{2017}{July}{}.
\newblock
{\BBOQ}\APACrefatitle {Gated Self-Matching Networks for Reading Comprehension
  and Question Answering} {Gated self-matching networks for reading
  comprehension and question answering}.{\BBCQ}
\newblock
\BIn{} \APACrefbtitle {Proceedings of the 55th Annual Meeting of the
  Association for Computational Linguistics.} {Proceedings of the 55th annual
  meeting of the association for computational linguistics.}
\newblock
\APACaddressPublisher{Vancouver, Canada}{}.
\PrintBackRefs{\CurrentBib}

\bibitem [\protect \citeauthoryear {%
Xu%
\ \protect \BOthers {.}}{%
Xu%
\ \protect \BOthers {.}}{%
{\protect \APACyear {2016}}%
}]{%
xu2016improved}
\APACinsertmetastar {%
xu2016improved}%
\begin{APACrefauthors}%
Xu, Y.%
, Jia, R.%
, Mou, L.%
, Li, G.%
, Chen, Y.%
, Lu, Y.%
\BCBL {}\ \BBA {} Jin, Z.%
\end{APACrefauthors}%
\unskip\
\newblock
\APACrefYearMonthDay{2016}{December}{}.
\newblock
{\BBOQ}\APACrefatitle {Improved relation classification by deep recurrent
  neural networks with data augmentation} {Improved relation classification by
  deep recurrent neural networks with data augmentation}.{\BBCQ}
\newblock
\BIn{} \APACrefbtitle {Proceedings of COLING 2016, the 26th International
  Conference on Computational Linguistics: Technical Papers} {Proceedings of
  coling 2016, the 26th international conference on computational linguistics:
  Technical papers}\ (\BPGS\ 1461--1470).
\newblock
\APACaddressPublisher{Osaka, Japan}{}.
\PrintBackRefs{\CurrentBib}

\bibitem [\protect \citeauthoryear {%
Zeng%
, Liu%
, Lai%
, Zhou%
\BCBL {}\ \BBA {} Zhao%
}{%
Zeng%
\ \protect \BOthers {.}}{%
{\protect \APACyear {2014}}%
}]{%
zeng2014relation}
\APACinsertmetastar {%
zeng2014relation}%
\begin{APACrefauthors}%
Zeng, D.%
, Liu, K.%
, Lai, S.%
, Zhou, G.%
\BCBL {}\ \BBA {} Zhao, J.%
\end{APACrefauthors}%
\unskip\
\newblock
\APACrefYearMonthDay{2014}{August}{}.
\newblock
{\BBOQ}\APACrefatitle {Relation Classification via Convolutional Deep Neural
  Network.} {Relation classification via convolutional deep neural
  network.}{\BBCQ}
\newblock
\BIn{} \APACrefbtitle {Proceedings of COLING 2014, the 25th International
  Conference on Computational Linguistics: Technical Papers} {Proceedings of
  coling 2014, the 25th international conference on computational linguistics:
  Technical papers}\ (\BPGS\ 2335--2344).
\newblock
\APACaddressPublisher{Dublin, Ireland}{}.
\PrintBackRefs{\CurrentBib}

\bibitem [\protect \citeauthoryear {%
Zheng%
\ \protect \BOthers {.}}{%
Zheng%
\ \protect \BOthers {.}}{%
{\protect \APACyear {2017}}%
}]{%
zheng2017joint}
\APACinsertmetastar {%
zheng2017joint}%
\begin{APACrefauthors}%
Zheng, S.%
, Wang, F.%
, Bao, H.%
, Hao, Y.%
, Zhou, P.%
\BCBL {}\ \BBA {} Xu, B.%
\end{APACrefauthors}%
\unskip\
\newblock
\APACrefYearMonthDay{2017}{July}{}.
\newblock
{\BBOQ}\APACrefatitle {Joint Extraction of Entities and Relations Based on a
  Novel Tagging Scheme} {Joint extraction of entities and relations based on a
  novel tagging scheme}.{\BBCQ}
\newblock
\BIn{} \APACrefbtitle {Proceedings of the 55th Annual Meeting of the
  Association for Computational Linguistics (Volume 1: Long Papers)}
  {Proceedings of the 55th annual meeting of the association for computational
  linguistics (volume 1: Long papers)}\ (\BPGS\ 1227--1236).
\newblock
\APACaddressPublisher{Vancouver, Canada}{Association for Computational
  Linguistics}.
\newblock
\begin{APACrefURL} \url{http://aclweb.org/anthology/P17-1113} \end{APACrefURL}
\PrintBackRefs{\CurrentBib}

\end{thebibliography}

\label{lastpage}

\end{document}